\documentclass[final,12pt]{elsarticle}
\usepackage[]{natbib}
\usepackage{amsfonts}
\usepackage{amsmath}
\usepackage{amssymb}
\usepackage{array}
\usepackage{balance}
\usepackage{bbding}
\usepackage{booktabs}
\usepackage{cite}
\usepackage{color}
\usepackage{graphicx}
\usepackage{hyperref}
\usepackage{lineno}
\usepackage{makecell}
\usepackage{multicol}
\usepackage{multirow}
\usepackage{stfloats}
\usepackage{textcomp}
\usepackage{verbatim}
\usepackage{url}
\usepackage{algorithm}
\usepackage{algpseudocode}
\journal{Elsevier}

\begin{document}
\begin{frontmatter}

\title{USRNet: Unified Scene Recovery Network for Enhancing Traffic Imaging under Multiple Adverse Weather Conditions}
\author[1]{Yuxu Lu}
\ead{yuxulouis.lu@connect.polyu.hk}
\author[2]{Ai Chen}
\ead{aichen@std.uestc.edu.cn}
\author[1]{Dong Yang}
\ead{dong.yang@polyu.edu.hk} 
\author[3]{Ryan Wen Liu}
\ead{wenliu@whut.edu.cn} 
\affiliation[1]{organization={Department of Logistics and Maritime Studies, The Hong Kong Polytechnic University},
	postcode={999077}, 
	state={Hong Kong}}
\affiliation[2]{organization={School of Computer Science and Engineering, University of Electronic Science and Technology of China},
	postcode={611731}, 
	state={Sichuan, China}}
\affiliation[3]{organization={School of Navigation, Wuhan University of Technology},
	postcode={430063}, 
	state={Wuhan, China}}
\begin{abstract}
    Advancements in computer vision technology have facilitated the extensive deployment of intelligent transportation systems and visual surveillance systems across various applications, including autonomous driving, public safety, and environmental monitoring. However, adverse weather conditions such as haze, rain, snow, and more complex mixed degradation can significantly degrade image quality. The degradation compromises the accuracy and reliability of these systems across various scenarios. To tackle the challenge of developing adaptable models for scene restoration, we introduce the unified scene recovery network (USRNet), capable of handling multiple types of image degradation. The USRNet features a sophisticated architecture consisting of a scene encoder, an attention-driven node independent learning mechanism (NILM), an edge decoder, and a scene restoration module. The scene encoder, powered by advanced residual blocks, extracts deep features from degraded images in a progressive manner, ensuring thorough encoding of degradation information. To enhance the USRNet's adaptability in diverse weather conditions, we introduce NILM, which enables the network to learn and respond to different scenarios with precision, thereby increasing its robustness. The edge decoder is designed to extract edge features with precision, which is essential for maintaining image sharpness. Experimental results demonstrate that USRNet surpasses existing methods in handling complex imaging degradations, thereby improving the accuracy and reliability of visual systems across diverse scenarios. The code resources for this work can be accessed in \textit{\url{https://github.com/LouisYxLu/USRNet}}.
\end{abstract}
%
\begin{highlights}
    \item Unified Scene Restoration Network (USRNet) for weather-affected image quality.
    \item Modular design with global context modeling for enhanced image restoration.
    \item Attention-driven NILM for robust scene recovery in various weather.
    \item Hybrid loss function integrating L1, contrastive, and edge losses for effective training.
    \item Superior performance in handling complex image degradations compared to existing methods.
\end{highlights}
\begin{keyword}
    Intelligent visual systems
    \sep Image degradation 
    \sep Adverse weather
    \sep Unified scene restoration
    \sep Neural network
\end{keyword}
\end{frontmatter}
%
%
\section{Introduction}\label{sec:introd}
   With the rapid advancement of artificial intelligence (AI) and computer vision (CV), vision-driven intelligent transportation systems (VITS), such as intelligent vehicles and surveillance, have become essential to contemporary society \citep{wan2022edge}. However, unexpected weather environmental factors such as haze, rain, and snow can significantly degrade imaging quality, thereby adversely impacting the accuracy and reliability of VITS \citep{liu2023aioenet}. For example, unnatural imaging process can lead to decreased accuracy in vehicle detection, license plate recognition, and pedestrian detection, consequently increasing the risk of traffic accidents \citep{husain2020vehicle}. To address these challenges, researchers have extensively proposed methods for restoring degraded images, which can be mainly categorized into two classes: single- and multi-scene degradation restoration. Single-scene methods focuse on addressing image quality issues caused by a specific type of environmental factor, such as dehazing \citep{he2010single,chen2024dea}, deraining \citep{fu2017clearing}, and desnowing \citep{liu2018desnownet}. In contrast, multi-scene methods are mainly used to handle various types of degradations \citep{liu2023aioenet,zhu2023learning,liu2024residual}, or even mixed degradations (e.g. rain mixed with haze) \citep{xu2024mvksr,guo2024onerestore}, simultaneously.
    \begin{figure}[t]
        \centering
        \setlength{\abovecaptionskip}{0.cm}
        \includegraphics[width=0.75\linewidth]{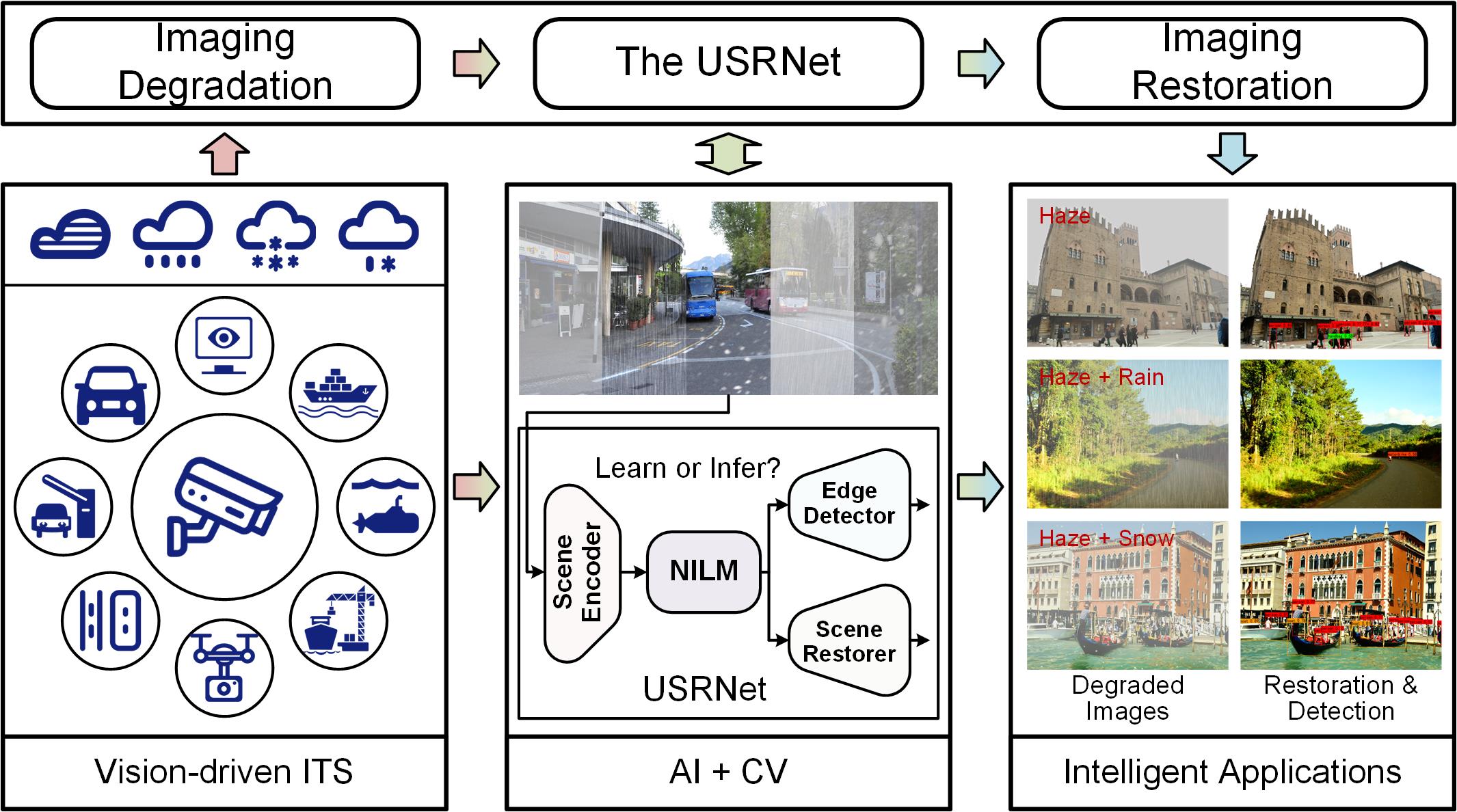}
        \caption{Severe weather conditions can significantly compromise the performance of VITS, resulting in decreased traffic efficiency and safety risks. Notwithstanding the propensity for weather-related disruptions to compromise transportation imaging sensors, the integration of AI and CV capabilities can effectively alleviate these disturbances, thereby ensuring an efficient and secure transportation system that can operate optimally even in the most adverse environmental conditions.}
        \label{Figure_Application}
    \end{figure}
    \begin{figure*}[t]
        \centering
        \setlength{\abovecaptionskip}{0.cm}
        \includegraphics[width=1.00\linewidth]{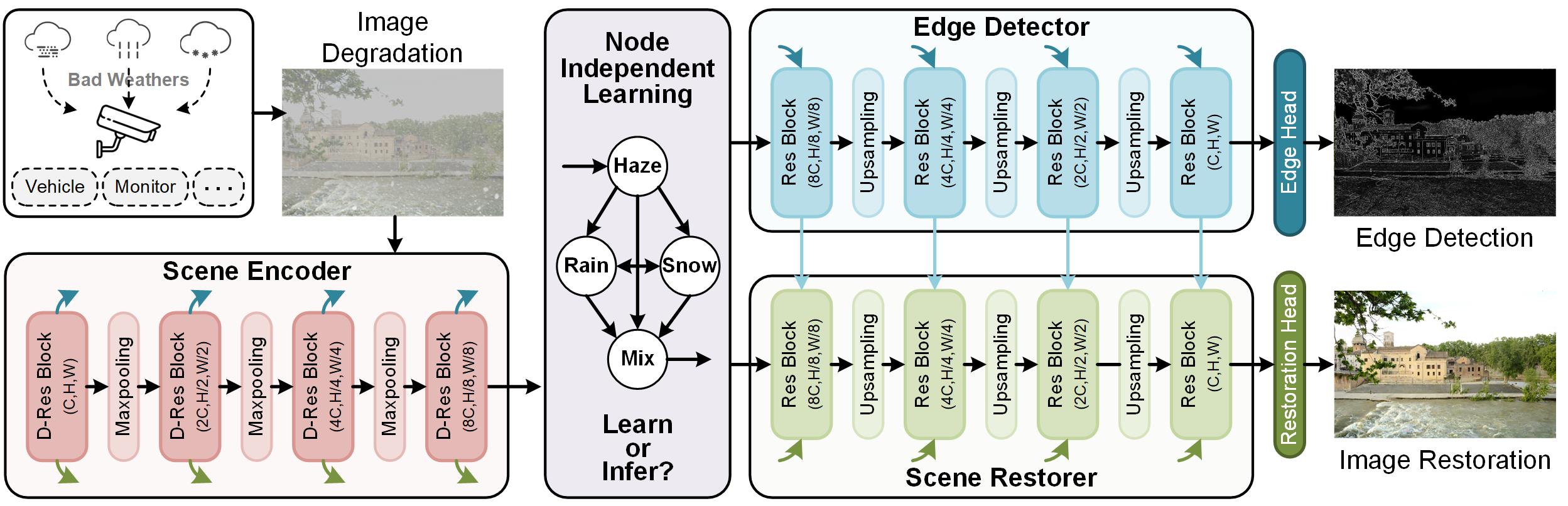}
        \caption{Overview of the proposed USRNet for image restoration under complex imaging conditions, demonstrated through edge detection and image restoration tasks. The scene encoder extracts multi-scale generic visual representations from the degraded image. NILM incorporates a dedicated training node for each type of degradation, enabling each node to learn more specific and focused features, thereby enhancing the overall restoration performance. The edge decoder generates potential edge features, assisting the scene restorer in producing the final restored image.}
        \label{Figure_Flowchart}
    \end{figure*}

    Image restoration methods have witnessed a remarkable evolution, transitioning from traditional to learning-based methods. Traditional dehazing methods, including the classic dark channel prior (DCP)-based \citep{he2010single} and Retinex-based \citep{fu2014retinex} methods, often struggle with bright objects or sky regions and with varying densities and complex structures. However, the emergence of learning-based methods like MSCNN \citep{ren2016single} and DehazeFormer \citep{song2023vision} has significantly improved dehazing performance by leveraging convolutional neural networks (CNN) and Transformer training. Recent advancements in dehazing have further integrated self-paced semi-curricular learning strategies \citep{guo2023scanet}, attention mechanisms \citep{chen2024dea}, and contrastive learning \citep{liu2024dfp} to boost performance. For deraining and desnowing tasks, traditional methods such as guided filtering- \citep{xu2012removing} and hierarchical-based \citep{wang2017hierarchical} methods have limitations in handling complex scenes. Learning-based methods, such as DerainNet \citep{fu2017clearing} and DesnowNet \citep{liu2018desnownet}, have significantly improved the effectiveness and efficiency of deraining and desnowing tasks. Moreover, other learning methods like invertible neural networks \citep{quan2023image} and generative adversarial networks (GAN) \citep{yang2022rain} have also been applied for different image restoration tasks. Despite the rapid development of single-scene restoration methods, the complex imaging environment has increased the demand with enhanced scene generalization ability. Multi-scene image restoration is a crucial field in computational imaging that aims to reverse the effects of image degradation caused by a multitude of environmental conditions, such as haze, rain, snow, and changes in illumination \citep{liu2023aioenet}. Hand-crafted filters and statistical models were used in traditional methods \citep{liu2022tape}, which were frequently customized to address particular categories of degradation. Learning-based methods mainly encompass GAN-based \citep{guo2020joint,cheng2023highway}, Transformer-based methods \citep{chen2021pre}, diffusion-based \citep{ozdenizci2023restoring,ye2024learning}, language-image-guided \citep{lin2024improving,guo2024onerestore}, etc. However, it is still challenging to extract the latent detail features when images suffer from mixed and complex degradation, demanding more accurate feature disentanglement and adaptive learning strategies.

    To address the challenge of creating adaptable models for scene restoration in VITS, we propose the unified scene recovery network (termed USRNet) that can handle multiple imaging degradation types. As shown in Fig. \ref{Figure_Flowchart}, USRNet is a sophisticated architecture comprising a scene encoder, an attention-driven node independent learning mechanism (NILM), an edge decoder, and a scene restorer. Specifically, the scene encoder, leveraging advanced residual blocks, progressively extracts deep features from degraded images, ensuring comprehensive degradation encoding. To enhance USRNet's versatility in varying weather scenarios, we innovate with NILM, enabling the network to independently learn and respond to different conditions with precision, thereby boosting its robustness. The edge detector can meticulously extract edge features, crucial for maintaining image sharpness. To optimize the network's performance across a broad spectrum of degradations, we devise a hybrid loss function that integrates multiple loss components, finely tuning the training to capture diverse degradation nuances. Extensive experimental results validate USRNet's superiority, showcasing its exceptional capability in handling complex image degradation scenarios. The main contributions of this work can be summarized as follows
    \begin{itemize}
        \item  We propose a novel unified scene recovery network (USRNet) designed to effectively address various types of degradations or even mixed degradations, significantly improving image restoration in challenging adverse weather conditions in VITS.

        \item  The proposed independent learning units of NILM can conduct targeted learning on the degradation features under different weather conditions during the training phase, and can jointly restore complex mixed degraded images during the inference phase while avoiding over-restoration of a single type of degradation.

        \item  Extensive experiments demonstrate USRNet's robust performance in handling diverse degradations, surpassing existing restoration methods. Furthermore, it has proven to be highly effective in object detection, which has significant application value in VITS.
    \end{itemize}    

    The rest of this paper is organized as follows. Section \ref{sec:relatedwork} provides a comprehensive review of existing research on multi-scene recovery. Section \ref{sec:pf} is problem formulation of degraded imaging model. The USRNet architecture is meticulously detailed in Section \ref{sec:usrnet}, outlining its innovative design. Performance assessments of USRNet, including experimental results and analysis, are presented in Section \ref{sec:experiments}, demonstrating its efficacy in VITS. Finally, Section \ref{sec:conclusion} concludes our work.

\section{Related Work}\label{sec:relatedwork}
    In this section, we will review related image restoration work from two aspects: single-scene and multi-scene.

\subsection{Single-Scene Restoration}
\subsubsection{Image Dehazing}
    Dehazing methods have evolved significantly, transitioning from traditional image processing algorithms to learning-driven models. He \textit{et al.} \citep{he2010single} proposed the DCP, which inverts the atmospheric scattering model to produce haze-free images by uncovering the statistical properties of hazy images. However, DCP-based methods \citep{fattal2014dehazing,liu2022rank} struggle with images containing bright objects or sky regions and are often ineffective in handling non-uniform haze due to the assumption of a uniform haze layer. Retinex-based methods \citep{fu2014retinex,kandhway2023adaptive} focus on filtering degraded images to retrieve light images but typically fail to accurately remove haze in regions with varying densities and complex structures. Learning-based methods \citep{zhu2023spectral,chen2024dea} have seen data-driven methods take the forefront in recent years. For example, SDCE \citep{zhu2023spectral} utilizing spectral dual-channel encoding to separately process high- and low-frequency image components, significantly enhancing dehazing performance by improving high-frequency details. Chen \textit{et al.} \citep{chen2024dea} proposed a detail-enhanced attention module that combines detail-enhanced convolution and content-guided attention to improve feature learning and dehazing performance. Unsupervised contrastive learning and adversarial training methods \citep{liu2024dfp} have also been employed to leverage unpaired real-world hazy and clear images. These methods represent significant advancements in the field, contributing to more robust and effective restoration in diverse real-world scenarios.
    \begin{figure}[t]
        \centering
        \setlength{\abovecaptionskip}{0.cm}
        \includegraphics[width=0.75\linewidth]{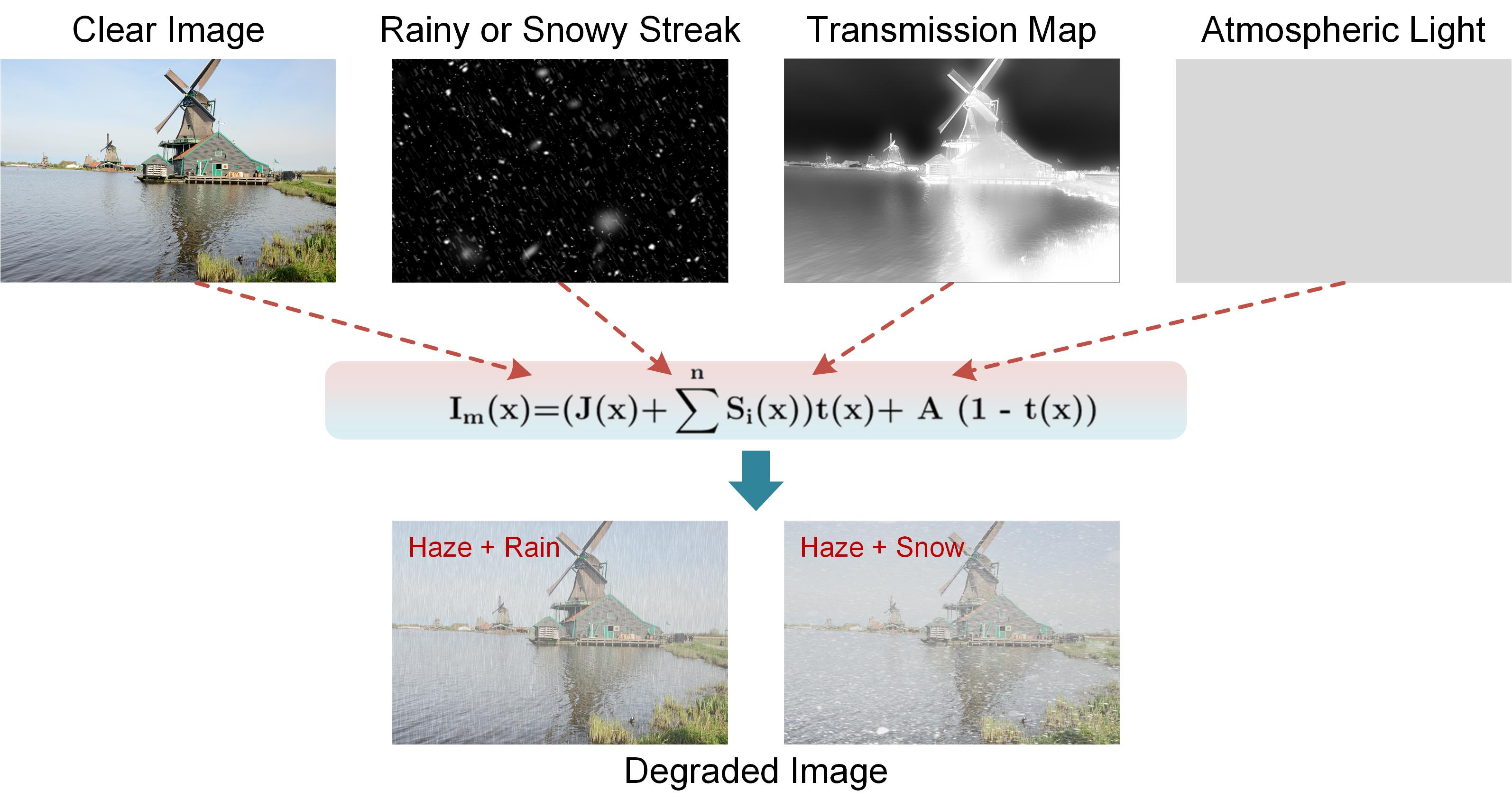}
        \caption{Illustration of the imaging degradation model under complex weather conditions, where uncertain combinations of factors yield a diverse range of degraded images.}
        \label{Figure_ImagingModel}
    \end{figure}
\subsubsection{Image Deraining}
    Earlier studies relied on filtering and statistics-based methods, for example, guided filters \citep{xu2012removing} have been effective in removing raindrops and snowflakes from images without relying on pixel-by-pixel statistical information. Hierarchical-based methods employing image decomposition and dictionary learning \citep{wang2017hierarchical} have also been used to efficiently remove raindrops and snowflakes from monochrome images through a multilayered strategy. However, these methods exhibit limited performance when dealing with complex and variable rain and snow scenes. CNN-based networks like DerainNet\citep{fu2017clearing} was among the early attempts to handle unwanted rain streak in images. Zhao \textit{et al.} \citep{zhao2024cycle} recently proposed a recurrent contrastive adversarial learning method with structural consistency to enhance the quality of rain removal images. Yang \textit{et al.} \citep{yang2024single} utilized a multiscale hybrid fusion network to merge multiscale features, which uses a non-local fusion module and an attention fusion module to generate superior rain-free images. The conditional GAN \citep{yang2022rain} for single-image rain removal and enhanced the outcomes by incorporating adversarial loss. GAN-based methods have significantly advanced the handling of complex rainy scenes, as well as the quality and efficiency of image restoration.
\subsubsection{Image Desnowing}
    Early image desnowing methods relies on smoothing filters, as proposed by He \textit{et al.} \citep{xu2012removing} utilized these filters to remove snow from single images. With the progression towards more complex algorithms, Wang \textit{et al.} \citep{wang2017hierarchical} proposed a three-layer hierarchical scheme combining image decomposition and dictionary learning to tackle both rain and snow. DesnowNet \citep{liu2018desnownet} utilizes multiple scales to model the diversity of snow and effectively remove opaque snow particles. Furthermore, Zhang \textit{et al.} \citep{zhang2021deep} proposed a deep-density multiscale network, which leverages deep prior and semantic information for image snow removal through a self-attention mechanism. Quan \textit{et al.} \citep{quan2023image} utilized inverse neural networks for single-image snow removal, achieving precise snow removal while preserving image details. Additionally, the JSTASR \citep{chen2020jstasr} was designed to classify and further remove snow. Wavelet transform and contradictory channel features were proposed by Chen \textit{et al.} \citep{chen2021all} to remove snow hierarchically, using a dual-tree complex wavelet representation.
\subsection{Multi-Scene Restoration}
    \begin{figure*}[t]
        \centering
        \setlength{\abovecaptionskip}{0.cm}
        \includegraphics[width=0.925\linewidth]{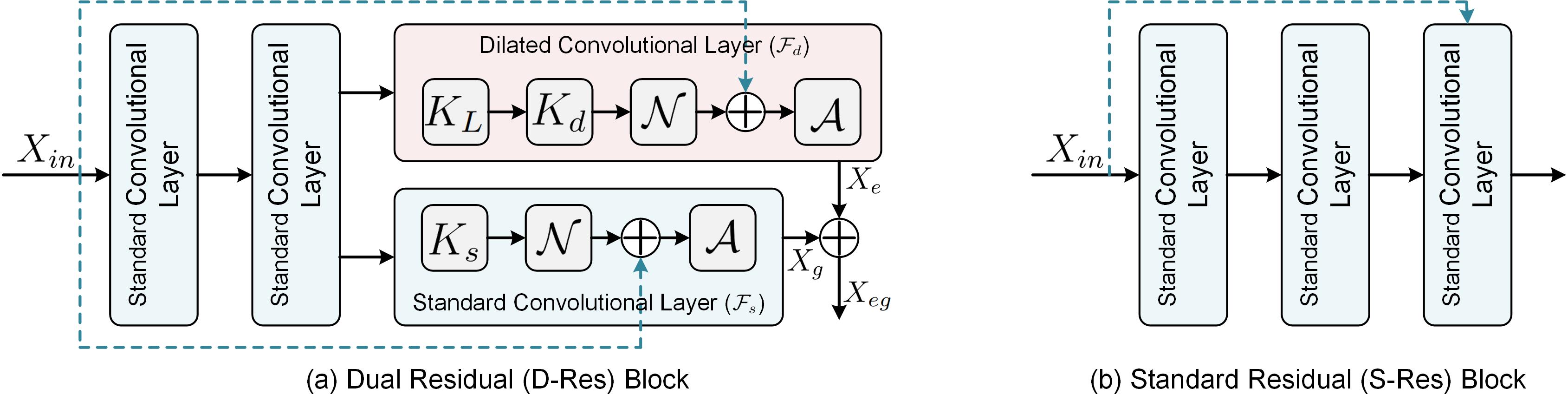}
        \caption{The pipeline of proposed dual residual (D-Res) block and standard residual (S-Res) block. D-Res will provide two output heads, each dedicated to learning and reasoning about edge features and global features of degraded images, respectively.}
        \label{Figure_Res}
    \end{figure*}
    Multi-scene image restoration is proposed to restore and enhance degraded images under various environmental conditions, including but not limited to haze, rain, and snow. Traditional learning methods are often challenging to address multiple types of image degradation using a single model. To this end, Guo \textit{et al.} \citep{gou2020clearer} proposed multiscale neural architecture search to optimize image restoration by incorporating parallel, transitional, and fusion operations. The multi-stage progressive restoration framework \citep{zamir2021multi} is proposed to enhance image quality through successive enhancement stages. Architectural search \citep{li2020all} has also been used to develop unified models to mitigate the effects of severe weather. Li \textit{et al.} \citep{li2022all} proposed a novel framework capable of addressing unknown types of image degradation. Chen \textit{et al.} \citep{chen2021pre} leveraged pre-trained Transformer models for a range of image processing tasks, demonstrating significant performance enhancements. Patil \textit{et al.} \citep{patil2023multi} proposed a domain translation-based unified method to achieve robust restoration to complex weather conditions by learning features of multiple weather degradations. The TAENet \citep{fang2024taenet} integrated a cross-encoder architecture and depth perception to enhance image quality, while MvKSR \citep{xu2024mvksr} used multiview knowledge-guided filtering to address challenges posed by haze and rain. Liu \textit{et al.} \citep{liu2024real} proposed the ERANet, which can adapt to the enhancement of images under different weather conditions by combining edge reparameterization and attention mechanism, achieving high-quality image restoration and low computational cost. Gao \textit{et al.} \citep{gao2024prompt} developed a cue-based, component-guided restoration method to handle multiple image degradation scenarios in a unified manner. The Uformer \citep{wang2022uformer} proposed a U-shaped Transformer architecture with locally enhanced window self-attention and multi-scale inpainting modulators to improve imaging quality. MPerceiver \citep{ai2024multimodal} used multimodal prompts to enhance adaptiveness, generalizability, and fidelity in various complex real-world scenarios. Ye \textit{et al.} \citep{ye2024learning} proposed a diffusion texture prior model that explicitly models high-quality texture details and incorporates conditional guidance adapters to achieve high fidelity and realistic textures in image restoration tasks, outperforming existing methods. Although current methods can handle concurrent multiple types of degradation, the nuanced effects of superimposing different types of degradation have not been thoroughly investigated, especially in the VITS.
\section{Problem Formulation}\label{sec:pf}
\subsection{Atmospheric Scattering Model}
    The atmospheric scattering model (ASM) is frequently used to describe how atmospheric particles, such as haze, smoke, and dust, scatter and absorb light, thereby causing image degradation. The ASM can be expressed as
    \begin{equation}   
        I_h(x) = I(x) t(x) + A (1 - t(x)),
    \end{equation}    
    where $I_h(x)$ is the pixel value of the hazy image, $I(x)$ is the pixel value of the haze-free image, $A$ is the atmospheric light (often approximated as the brightest pixel value), and $t(x)$ is the transmission map, representing the proportion of light that is not scattered as it travels from the scene object to the camera. The transmission map is typically defined as $t(x) = e^{-\beta d(x)}$, where $\beta$ is the atmospheric scattering coefficient and $d(x)$ is the distance between the object and the camera.
\subsection{Rain or Snow Model}
    Following the methodologies outlined in~\citep{li2019heavy, chen2021all}, our process superimposes rain or snow streaks $S(x)$ onto the clear images. Therefore, we can simply synthesize the rain- or snow-degraded image $I_{rs}(x)$, which is expressed as
    \begin{equation}\label{eq:haze}
        I_{rs}(x) = I(x) + S(x).
    \end{equation}

    Rain or snow streaks in the real world exhibit various shapes, resulting in an irregular distribution of streaks. To more accurately represent the appearance of streaks in degraded images, the rain- or snow-degraded model can be optimized to include multiple streak layers, that is,
    \begin{equation}\label{eq:rainsnow}
        I_{rs}(x)= I(x) +\sum^{n} S_{i}(x),
    \end{equation}
     where each $S_{i}$ is a layer of rainy or snowy streaks with the same direction, $i$ and $n$ are the index and maximum number of the streak layers.
\subsection{Mixed Degradation Model}
    The imaging model for rain or snow describes how images are affected by rainy or snowy steaks, including light scattering, absorption, and occlusion effects \citep{zhang2023data}. As shown in Fig. \ref{Figure_ImagingModel}, Combining Eq. \ref{eq:haze} and Eq. \ref{eq:rainsnow}, the mixed degradation image $I_m(x)$ can be expressed as
    \begin{equation}\label{eq:model}
        I_m(x) = (I(x)+\sum^{n} S_{i}(x)) t(x) + A (1 - t(x)).
    \end{equation}    

    In real-world applications, the imaging model's complexity increases to accommodate the dynamic features of raindrops or snowflakes, as well as the multiple scattering effects of light.

\section{USRNet: Unified Scene Recovery Network}\label{sec:usrnet}
    USRNet is architected with a modular method and incorporates global and edge context modeling to restore degraded images in multiple challenging weather scenarios. The architecture of the USRNet is explained in detail in this section, including the scene encoder, NILM, edge decoder, scene restorer, and loss function.
    \begin{figure*}[t]
        \centering
        \setlength{\abovecaptionskip}{0.cm}
        \includegraphics[width=0.925\linewidth]{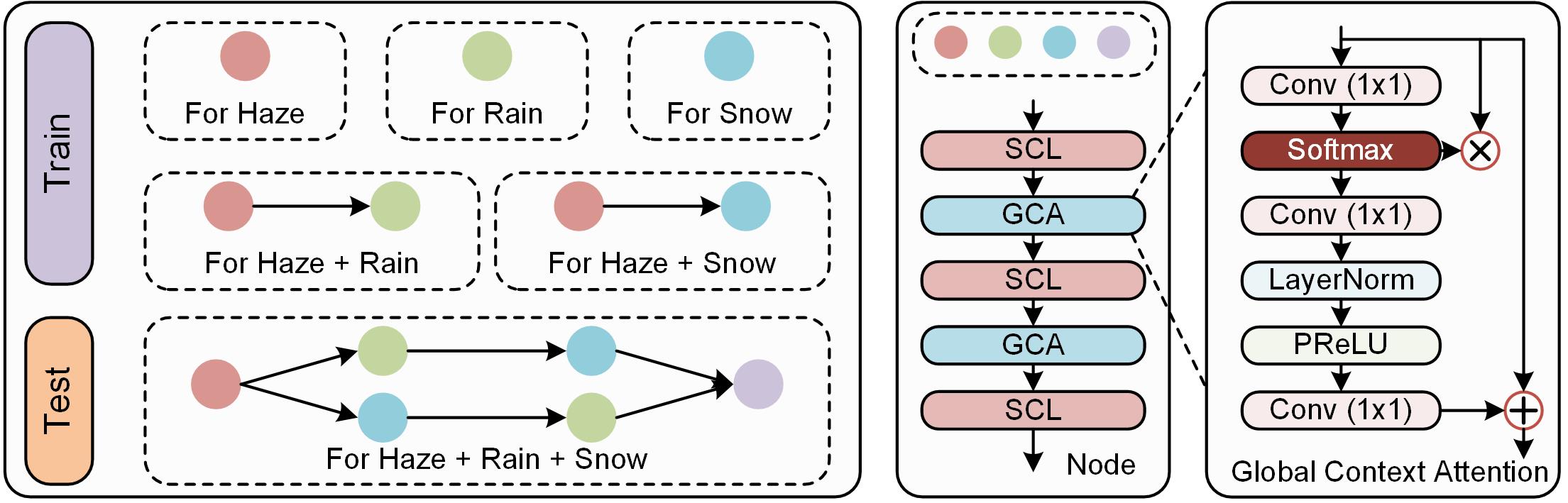}
        \caption{The pipeline of proposed NILM. The standard convolutional layer (SCL) and global context attention (GCA) are used to extract long-range dependencies and global context information. During inference phase, NILM can generate latent features by adaptively calling parameters to specific training nodes for each type of degradation.}
        \label{Figure_NILM}
    \end{figure*}
\subsection{Scene Encoder}
\subsubsection{Standard Convolutional Layer}
    The scene encoder effectively extracts features from degraded images, supplying rich semantic information for subsequent network modules. It is composed of four dual residual (D-Res) blocks, each designed to progressively extract multi-scale features through convolutional and max-pooling layers. For the input feature $X_{in}$, the simplified mathematical representation of a standard convolutional layer (SCL, $\mathcal{F}_s$) is
    \begin{equation}   
        X_{out}^s = \mathcal{F}_s(X_{in}) = \mathcal{A}(\mathcal{N}(K_s * X_{in} + b)),
    \end{equation}  
    where $X_{out}^s$ is the SCL-based output feature map, $K_s$ is the parameter-learnable standard convolution kernel, $b$ is the bias, $\mathcal{N}$ is the normalization operation, and $\mathcal{A}$ is the activation function. In this work, we suggest layer normalization and parametric rectified linear unit, respectively. 
\subsubsection{Frequency-Dependent Feature Extraction}
    We improve our method by incorporating dilated convolutions for edge feature encoding, which increases the model's capacity to accommodate edge information in light of the ambiguity surrounding the effects of different degradation types on image edges. Dilated convolution can enhance edge global features by expanding the receptive field, capturing broader contextual information, and addressing widespread degradation effects. In addition, we suggest to use the Laplacian operator as a non-learnable convolution kernel to calculate edge gradients of feature maps at different scales. Mathematically, for the input feature map $F(x,y)$, the Laplacian operator $\nabla^2 F$ can be expressed as
    \begin{equation}
        \nabla^2 F=\frac{\sigma^2 F}{\sigma x^2}+\frac{\sigma^2 F}{\sigma y^2},
    \end{equation}
    where $\frac{\partial^2 F}{\partial x^2}$ and $\frac{\partial^2 F}{\partial y^2}$ are the second-order partial derivative of $F$ with respect to the $x$ and $y$ direction. The Laplace operator is usually represented by a fixed convolution kernel $K_L$ in convolution form, that is,
    \begin{equation}\label{eq:lapkernel}
        K_L = \begin{bmatrix}
        0 & -1 & 0 \\
        -1 & +4 & -1 \\
        0 & -1 & 0
        \end{bmatrix}.
    \end{equation}

    Therefore, we can obtain high-frequency edge gradient features $X_{h}$, which can be given as
    \begin{equation}
        X_{h} = K_d * (K_L * X_{in}) + b,
    \end{equation}
   where $K_d$ is the parameter-learnable dilated convolution kernel. In this work, we subtract the high-frequency feature $X_{h}$ from the input feature map $X_{in}$ to obtain the low-frequency feature, thereby weakening the impact of unwanted rain and snow marks, i.e.,
    \begin{equation}
        X_{l} =  K_d * (X_{in} - K_L * X_{in}) + b,
    \end{equation}

    The high and low frequency features generated by the non-learnable operators can reduce the network's dependence on the training dataset, thereby improving its robustness in recovery under different degradations.
\subsubsection{Multi-view Feature Fusion}
    High-frequency information is crucial for preserving fine details and textures, low-frequency information is essential for maintaining the overall shape and structure, and normal convolutional features provide a balance between the two. Therefore, we will three types features to jointly optimize the network, i.e.,
    \begin{equation}   
        X_{out}^d = \mathcal{A}(\mathcal{N}(K_s * (X_{h} + X_{l} + X_{n}) + b)),
    \end{equation}  
    where $X_{out}^d$ is the output feature map. The scene encoder generates encoding features that are subsequently provided to both the edge decoder and the scene restorer. The suggested D-Res can leverage global features to provide overall structure and background information while using edge features to enhance sharpness and detail.
\subsection{Node Independent Learning Mechanism}
\subsubsection{Principle of NILM}
    To further enhance restoration performance, as shown in Fig. \ref{Figure_NILM}, in the training phase, the proposed NILM (assuming $\psi_N$) provides dedicated training nodes for each degradation type, thereby focusing on specific degradation types and learning more specific features to enhance the overall restoration effect. Let $D_i$ represent the degraded image of the $i$-th degradation type, and $\psi_{N_i}$ represent the dedicated learning node trained for $D_i$. Therefore, the training process can be expressed as
    \begin{equation}
        \psi_{N_i}=\operatorname{Train}\left(\psi_N, D_i\right).
    \end{equation}

    In the testing phase, all nodes will be called sequentially according to the imaging model mentioned in literature \citep{guo2024onerestore}, so as to be more robust to various random image restoration requirements. Therefore, for the input feature of NILM (i.e., $F_{enc}$), its restoration process can be expressed as
    \begin{equation}
         F_{\text{NILM}}=\psi_{N_1}\left(\psi_{N_2}\left(\ldots \psi_{N_n}(F_{\text{enc}}) \ldots\right)\right),
    \end{equation}
    where $\psi_{N_1}$, $\psi_{N_2}$, $\ldots$, and $\psi_{N_n}$ are dedicated sub-model nodes obtained during the training phase. 

    NILM enables the model to gradually process different types of degradation in the image, thereby restoring the image quality more comprehensively. Each sub-model node focuses on processing a specific type of degradation, so throughout the restoration process, the model can better meet with various random image restoration needs, improving the robustness and adaptability of the model.
\subsubsection{Network Structure of NILM}
    The NILM combines SCL and GCA to extract long-range dependencies and global context information. SCL is mainly used to extract local features by leveraging local receptive fields for feature extraction, ensuring that each position’s features extract high-frequency information and detailed structures within its neighborhood. GCA is used to ensure the richness and completeness of local features, especially when dealing with complex image details such as raindrops, snowflakes, and haze. Therefore, for NILM, the input features are $X_n \in \mathbb{R}^{H \times W \times C}$, where $H$, $W$, and $C$ denote the height, width and number of channels of the feature map, respectively. Therefore, we first perform global pooling on the input feature $\mathbf{X}$ to obtain the global feature vector $\mathbf{z} \in \mathbb{R}^{C}$, which can be given as
    \begin{equation}   
        \mathbf{z} = \frac{1}{H \times W} \sum_{i=1}^{H} \sum_{j=1}^{W} X_n^{i,j},
    \end{equation}   
   where $X_n^{i,j}$ represents the feature vector at position $(i, j)$. Next, we apply a series of transformations to the global feature vector $\mathbf{z}$ to obtain a new feature vector $\mathbf{z}'$, which are typically implemented using a fully connected layer ($\mathcal{F}_{fcl}$). Finally, we integrate the transformed global features $\mathbf{z}'$ into the local features at each position $X_n^{i,j}$, obtaining the attention-enhanced features $\mathbf{X}_{gca}^{i,j}$, i.e.,
    \begin{equation}  
        X_{gca}^{i,j} = X_n^{i,j} + \mathcal{F}_{fcl}(\mathbf{z}) = X_n^{i,j} + \mathbf{z}'.
    \end{equation}   

    The GCA effectively can extrace richer image features, significantly improving the restoration of image details but also ensure structural consistency and visual credibility. The effect of GCA was verified in subsection \ref{as}.
\subsection{Edge Detector}
    \setlength{\tabcolsep}{1.50pt}
    \begin{table}[t]
    	\centering
            \scriptsize
    	\caption{The details of training and testing datasets used in our experiment.}  
        \begin{tabular}{l|cc|ccc|cc}
        \hline
        {Datasets} & {Train} & {Test} & {Haze}         & {Rain}         & {Snow}         & Haze+Rain                     & Haze+Snow            \\\hline\hline
        RESIDE \citep{li2018benchmarking}                & 1000                   & 100                   & \CheckmarkBold &  &  &  &  \\
        Rain 100L \citep{yang2019joint}    & 1000    & 100     &    & \CheckmarkBold    &  &   &    \\
        CSD \citep{chen2021all}     & 1000       & 100    &    &    &     \CheckmarkBold   &    &    \\
        CDD-11 \citep{guo2024onerestore}    &     5915   & 1000    & \CheckmarkBold & \CheckmarkBold & \CheckmarkBold & \CheckmarkBold & \CheckmarkBold  \\\hline
        \end{tabular}\label{table_dataset}
    \end{table}
\subsubsection{Principle}
    Edge features are critical in images, defining the contours and structures of objects and containing a significant amount of high-frequency information. However, these features are easily damaged or blurred under adverse weather conditions such as rain, snow, and haze. The edge decoder addresses this by extracting these crucial edge features through multiple standard residual (S-Res, shown in Fig. \ref{Figure_Res} (b)) blocks with different resolution scales. It then restores the spatial dimensions of the feature maps via upsampling operations to match the input feature maps. It can ensure that the edge features are synchronized with those output by other network modules, facilitating subsequent fusion and processing. Ultimately, the feature maps produced by the edge decoder are rich in edge information, aiding the overall image restoration task and improving the quality of the restored images. The edge decoder can adaptively handle various complex weather scenarios. By combining the features output by other network modules, it provides more detailed information, significantly enhancing the overall effect and quality of image restoration.
    \begin{figure}[t]
        \centering
        \includegraphics[width=0.75\linewidth]{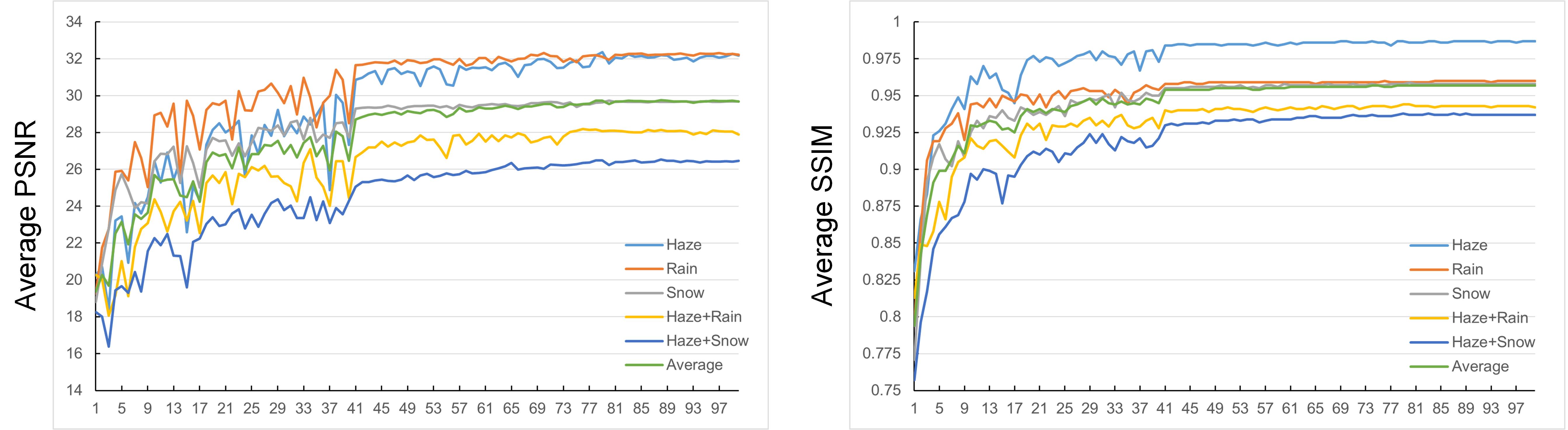}
        \caption{The convergence analysis under different degradation scenarios. The restoration performance of USRNet for different types of degradation tasks reaches a stable level at $\text{epoch}  = 80$.}
        \label{Figure_epoch}
    \end{figure}
\subsubsection{Loss Function}
    Laplacian edge loss is employed in image generation models to enhance the clarity and quality of generated images by emphasizing their edges. The process begins with the application of the Laplacian operator, a $3 \times 3$ convolution kernel (i.e., Eq. \ref{eq:lapkernel}) suggested for edge detection, to both the generated image $I_{\text{output}}$ and the target image $I_{\text{target}}$. The Laplacian edge loss is then calculated as the mean absolute error (MAE) of the absolute differences between these transformed images. Mathematically, edge loss $\mathcal{L}_{e}$ can be expressed as
    \begin{equation}  
        \mathcal{L}_{e} = \frac{1}{N} \sum_{i=1}^{N}  \left\| \textbf{Lap}(I_r) - \textbf{Lap}(I_t) \right\|_1,
    \end{equation}  
    where $\textbf{Lap}(\cdot)$ represents the Laplacian edge gradient feature extraction operation. The edge loss function measures the discrepancy between the edges of the generated and target images by comparing their edge information, which is obtained through the Laplacian transformation. It sums the absolute differences across all pixel locations. This process encourages the model to produce images with sharper and more defined edges, thereby enhancing the overall visual quality of the generated images.

\subsection{Scene Restorer}
\subsubsection{Principle}

    The scene restorer integrates features from the scene encoder, edge detector, and NILM, starting by aggregating these features to generate a feature map that encompasses both local and global information, that is,
    \begin{equation}  
       F_{\text{agg}} = F_{\text{enc}} \oplus F_{\text{edge}} \oplus F_{\text{NILM}},
    \end{equation}   
    where $F_{\text{enc}}$ represents the deep features extracted by the scene encoder, $F_{\text{edge}}$ represents the edge-preserving information from the edge detector, and $F_{\text{NILM}}$ encompasses the non-local features captured by NILM. Feature fusion ensures that the scene restorer has access to rich semantic feature information, including edge structure, color, contrast, and other essential elements for high-quality reconstruction. Similar to the edge detector's network structure, the scene restorer employs a multi-step process to fully leverage the extracted features, resulting in superior restoration performance across various degradation conditions. By carefully integrating and enhancing elements from multiple stages, the scene restorer guarantees a final output that is both intricately detailed and visually coherent, thereby achieving high-quality scene reconstruction.
\subsubsection{Loss Function}
    To train each sub node $\mathcal{N}_i$, we need to define a loss function $\mathcal{L}_i$, which measures the difference between the model output and the real image. We define each dataset $D_i$ as consisting of a set of degraded images $I_i^j$ and corresponding target images $I_t^j$, i.e., $D_i=\{(I_i^j, I_{t}^j)\}$. In this work, we still suggest to use the MAE loss function, then for the $i$-th sub model, the loss function can be expressed as
    \begin{equation}
        \mathcal{L}_{\text{MAE}}=\frac{1}{\left|D_i\right|} \sum_{\left(I_i^j, I_{t}^j\right) \in D_i}\left\|\mathcal{F}_{\mathcal{N}_i}\left(I_i^j ; \theta_i ; \theta_s\right)-I_{t}^j\right\|_1,
    \end{equation}
    where $\left|D_i\right|$ is the number of samples in dataset $D_i$, $\theta_i$ is the parameter of each sub-model $\mathcal{N}_i$, and $\theta_s$ is the shared parameter of suggested scene encoder and scene restorer.

    Contrastive Loss \citep{wu2021contrastive} is used to optimize single image restoration networks by pulling the restored image closer to the clear image and pushing it away from the degraded image. We suggest the pre-trained VGG19 network to extract features $\phi$, the distances for each layer's features are calculated as 

    \begin{equation}   
        \left\{
        \begin{array}{l}
        d_{ap}^i = \left\| \phi_i(I_t) - \phi_i(I_r) \right\|_1 \\
        d_{an}^i = \left\| \phi_i(I_t) - \phi_i(I_i) \right\|_1,
        \end{array}
        \right.
    \end{equation}  
    where $\phi_i$ denotes the feature extraction of the $i$-th layer of the VGG19 network, $d_{ap}^i$ is the distance between the anchor image and the positive sample image at the $i$-th layer, $d_{an}^i$ is the distance between the anchor image and the negative sample image at the $i$-th layer. Therefore, the contrastive loss $\mathcal{L}_{c}$ is then defined as
    \begin{equation}  
        \mathcal{L}_{c} = \sum_{i} w_i \cdot \frac{d_{ap}^i}{d_{an}^i + \epsilon},
    \end{equation}  
     where $w_i$ are the weights for each layer (typically $[ \frac{1}{32}, \frac{1}{16}, \frac{1}{8}, \frac{1}{4}, 1 ]$), and $\epsilon$ is a very small constant to prevent division by zero (usually $1 \times 10^{-7}$). This definition aims to optimize the dehazing network performance by comparing the distance between the restored image $I^r$ and the clear image $I^t$ ($d_{ap}$) with the distance between the restored image $I_r$ and the hazy image $I_d$ ($d_{an}$).
    
    The goal of the training process is to minimize the loss function of each sub-model to find the optimal parameter $\theta_i^*$. The optimization process can be expressed as
    \begin{equation}
        \theta_i^*=\arg \min _{\theta_i} (\gamma_1\mathcal{L}_{\text{MAE}} + \gamma_2\mathcal{L}_{c}),
    \end{equation}
    where $\gamma_1$ and $\gamma_2$ represent the weights of two different loss functions. We experimentally determined that setting $\gamma_1 = 0.85$ and $\gamma_2 = 0.15$ yields the best restoration results.
    \setlength{\tabcolsep}{0.50pt}
    \begin{table}[t]
        \centering
        \scriptsize
        \caption{Comparison of dehazing quantitative results (mean$\pm$std) with referenced and no-referenced evaluation metrics on RESIDE \citep{li2018benchmarking}. The best results are in \textbf{bold}, and the second best are with \underline{underline}.}
        \begin{tabular}{l|cccc|cc}
        \hline
        & PSNR $\uparrow$ & SSIM $\uparrow$ & FSIM $\uparrow$ & VSI $\uparrow$ & NIQE $\downarrow$ & PIQE $\downarrow$ \\ \hline\hline
        DCP \citep{he2010single}          & 16.770$\pm$2.992 & 0.774$\pm$0.079 & 0.933$\pm$0.027 & 0.972$\pm$0.012 & 2.992$\pm$0.790 & 10.620$\pm$4.889 \\ 
        MSCNN \citep{ren2016single}        & 15.439$\pm$4.579 & 0.771$\pm$0.131 & 0.901$\pm$0.073 & 0.964$\pm$0.028 & 3.382$\pm$1.071 & 13.185$\pm$8.920 \\ 
        AODNet \citep{li2017aod}       & 15.990$\pm$3.383 & 0.748$\pm$0.138 & 0.841$\pm$0.088 & 0.953$\pm$0.034 & 3.477$\pm$1.097 & 14.788$\pm$9.487 \\ 
        FFANet \citep{qin2020ffa}       & 18.198$\pm$6.535 & 0.803$\pm$0.155 & 0.904$\pm$0.091 & 0.966$\pm$0.035 & 3.481$\pm$1.152 & 10.736$\pm$8.600 \\ 
        DehazeFormer \citep{song2023vision} & 20.201$\pm$3.914 & 0.858$\pm$0.093 & 0.944$\pm$0.046 & 0.980$\pm$0.020 & 3.184$\pm$0.826 & 9.553$\pm$5.375  \\ 
        AiOENet \citep{liu2023aioenet}      & 22.712$\pm$4.211 & 0.899$\pm$0.072 & 0.964$\pm$0.029 & 0.987$\pm$0.012 & 3.367$\pm$0.737 & 10.336$\pm$5.623 \\ 
        AirNet \citep{li2022all}       & 16.509$\pm$5.369 & 0.738$\pm$0.144 & 0.880$\pm$0.078 & 0.955$\pm$0.037 & 3.747$\pm$1.008 & 7.684$\pm$4.708  \\ 
        TransW \citep{valanarasu2022transweather} & 19.001$\pm$5.413 & 0.847$\pm$0.107 & 0.941$\pm$0.049 & 0.977$\pm$0.023 & 3.761$\pm$0.798 & \textbf{5.810$\pm$3.569}  \\ 
        WeatherDiff \citep{ozdenizci2023restoring}  & 15.389$\pm$3.813 & 0.771$\pm$0.113 & 0.914$\pm$0.053 & 0.965$\pm$0.026 & 3.620$\pm$0.854 & 8.289$\pm$4.494  \\ 
        WGWSNet \citep{zhu2023learning}       & 17.265$\pm$5.717 & 0.829$\pm$0.125 & 0.942$\pm$0.058 & 0.975$\pm$0.027 & 3.103$\pm$0.808 & 9.865$\pm$5.605  \\ 
        MvkSR \citep{xu2024mvksr}        & \underline{23.159$\pm$4.553} & \underline{0.900$\pm$0.081} & \underline{0.973$\pm$0.025} & \underline{0.989$\pm$0.012} & \underline{2.991$\pm$0.806} & 10.253$\pm$4.642 \\ \hline
        USRNet       & \textbf{24.739$\pm$4.720}	&\textbf{0.907$\pm$0.080}	&\textbf{0.974$\pm$0.026}	&\textbf{0.990$\pm$0.012}	&\textbf{2.987$\pm$0.780} 	&\underline{6.813$\pm$3.519}                   \\ \hline
        \end{tabular}\label{table:dehazing}
    \end{table}
    \begin{figure}[t]
        \centering
        \setlength{\abovecaptionskip}{0.cm}
        \includegraphics[width=1.00\linewidth]{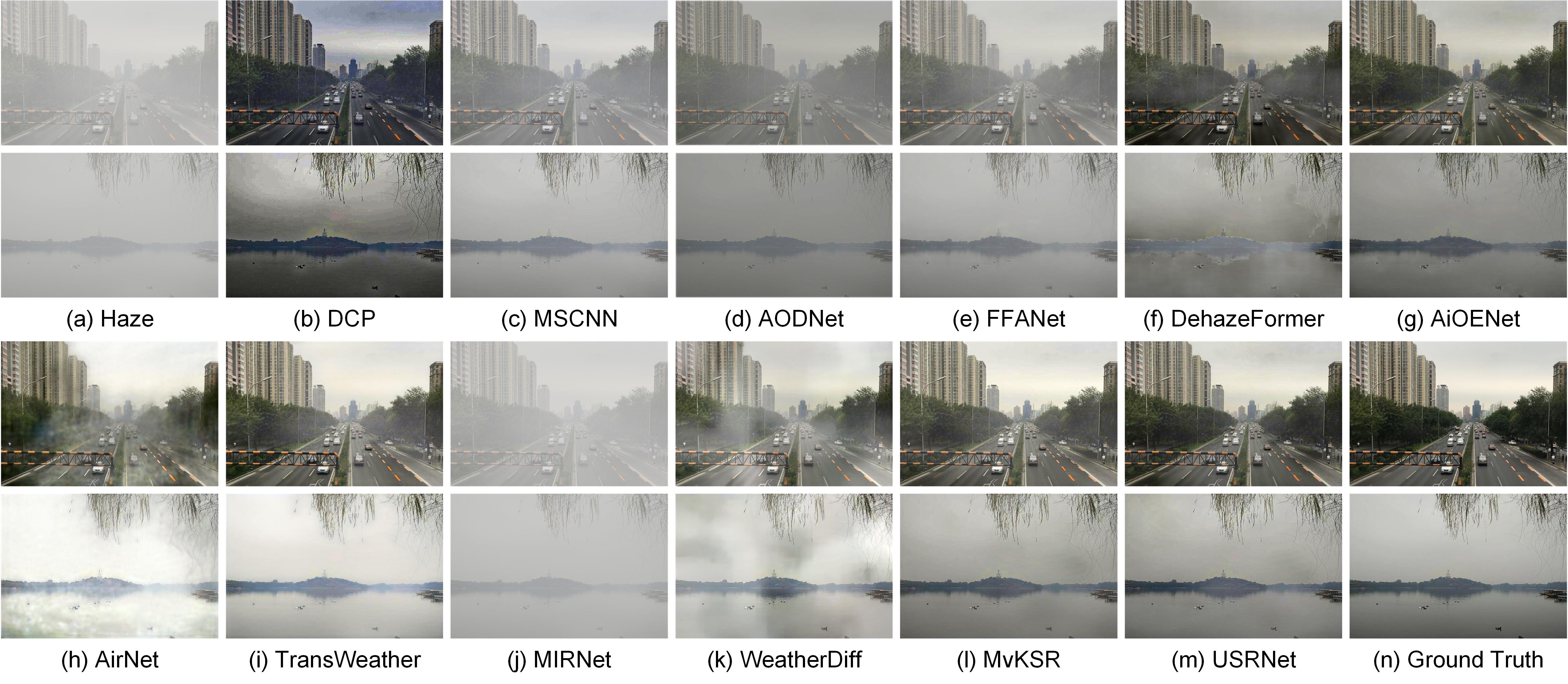}
        \caption{Visual comparisons of hazy scene recovery from RESIDE \citep{li2018benchmarking}. (a) Haze, restored images, generated by (b) DCP \citep{he2010single}, (c) MSCNN \citep{ren2016single}, (d) AODNet \citep{li2017aod}, (e) FFANet \citep{qin2020ffa}, (f) DehazeFormer \citep{song2023vision}, (g) AiOENet \citep{liu2023aioenet}, (h) AirNet \citep{li2022all}, (i) TransW \citep{valanarasu2022transweather}, (j) MIRNet \citep{zamir2022learning}, (k) WeatherDiff \citep{ozdenizci2023restoring}, (l) MvkSR \citep{xu2024mvksr}, (m) USRNet, and (n) Ground Truth, respectively.}
        \label{Figure_Dehazing}
    \end{figure}
\section{Experiments and Discussion}\label{sec:experiments}
    In this section, we provide a detailed overview of the experimental setup, including the train and test datasets, experimental platform, evaluation metrics, and competitive methods. We then present a comparison of USRNet with state-of-the-art methods on both standard and VITS-related datasets, providing quantitative and qualitative results that demonstrate its superiority. We conduct a series of ablation experiments to validate the design of the network and provide insights into its key components. Furthermore, we discussed the application of USRNet in advanced vision tasks and its running time and computational complexity.
\subsection{Implementation Details}
\subsubsection{Datasets and Experimental Platform}
    The scarcity of real-world paired data  (i.e., clear and low-visibility) complicates the training of learning-based image restoration networks. To address this issue, we leverage the realistic single-image dehazing (RESIDE) \citep{li2018benchmarking}, Rain100L \citep{yang2019joint} (including rain marks), comprehensive snow dataset \citep{chen2021all} (CSD, including snow marks) \citep{guo2024onerestore} to synthesize low-visibility images. Additionally, to enhance the generalization capabilities across diverse scenarios, we incorporate the composite degradation dataset (CDD-11) \citep{guo2024onerestore} into our training set. To evaluate the robustness and generalization performance of our proposed method, we conduct experiments using several standard datasets. These datasets include RESIDE \citep{li2018benchmarking} for dehazing, Rain100L for deraining, CSD for desnowing, and CDD-11 for multi-scene degradation (including haze, rain, snow, haze + rain, haze + snow). More detailed information regarding the datasets used for training and testing our USRNet is presented in Table \ref{table_dataset}. We train the network for 100 epochs using the Adam optimizer with an initial learning rate of 0.001. The learning rate is decayed by a factor of 0.1 every 40 epochs. The convergence curve of the learning phase is shown in Fig. \ref{Figure_epoch}. All experiments are conducted in a Python 3.9 environment using the PyTorch software package, leveraging a PC equipped with an Intel(R) Core(TM) i9-12900K CPU @ 2.30GHz and an Nvidia GeForce RTX 4090 GPU for accelerated computations. Our USRNet takes about 24 hours to complete training. During the inference phase, for an image with a resolution of 1080p (i.e., $1920 \times 1080$), it only takes 0.01s to restore image, which can meet the needs of real-time image restoration in VITS.
    \setlength{\tabcolsep}{0.50pt}
    \begin{table}[t]
        \centering
        \scriptsize
        \caption{Comparison of deraining quantitative results (mean$\pm$std) with referenced and no-referenced evaluation metrics on Rain100L \citep{yang2019joint}. The best results are in \textbf{bold}, and the second best are with \underline{underline}.}
        \begin{tabular}{l|cccc|cc}
        \hline
        & PSNR $\uparrow$ & SSIM $\uparrow$ & FSIM $\uparrow$ & VSI $\uparrow$ & NIQE $\downarrow$ & PIQE $\downarrow$\\ \hline\hline
        DDN \citep{fu2017removing}              & 27.047$\pm$2.745 & 0.848$\pm$0.073 & 0.917$\pm$0.046 & 0.978$\pm$0.014 & 4.022$\pm$1.064 & 7.266$\pm$4.359   \\ 
        DID \citep{zhang2018density}              & 24.085$\pm$2.548 & 0.799$\pm$0.081 & 0.895$\pm$0.047 & 0.971$\pm$0.015 & 4.182$\pm$1.119 & 16.121$\pm$9.070 \\ 
        LPNet \citep{fu2019lightweight}            & 32.561$\pm$3.158 & 0.935$\pm$0.031 & 0.961$\pm$0.017 & 0.991$\pm$0.004 & \textbf{3.089$\pm$0.675} & 6.882$\pm$3.622   \\ 
        DIG \citep{ran2020single}             & 30.222$\pm$2.995 & 0.907$\pm$0.036 & 0.945$\pm$0.023 & 0.987$\pm$0.006 & 3.784$\pm$0.732 & \underline{6.637$\pm$3.959}   \\ 
        DualGCN \citep{fu2021rain}            & \textbf{34.961$\pm$3.032}  & \textbf{0.966$\pm$0.016}  & \textbf{0.976$\pm$0.009}  & \textbf{0.994$\pm$0.003}  & 3.400$\pm$0.836  & 8.585$\pm$5.398   \\ 
        AiOENet \citep{liu2023aioenet}          & 33.024$\pm$3.372 & 0.945$\pm$0.026 & 0.957$\pm$0.015 & 0.990$\pm$0.004 & 3.678$\pm$0.746 & 11.663$\pm$9.138  \\ 
        AirNet \citep{li2022all}           & 34.702$\pm$3.326 & 0.956$\pm$0.023 & 0.974$\pm$0.017 & 0.994$\pm$0.003 & \underline{3.215$\pm$0.710} & 6.783$\pm$3.553   \\ 
        TransW \citep{valanarasu2022transweather}     & 22.397$\pm$3.714 & 0.794$\pm$0.101 & 0.895$\pm$0.059 & 0.969$\pm$0.020 & 4.067$\pm$1.129 & \textbf{4.871$\pm$2.711}   \\ 
        WeatherDiff \citep{ozdenizci2023restoring} & 20.683$\pm$2.578 & 0.793$\pm$0.102 & 0.890$\pm$0.064 & 0.965$\pm$0.023 & 3.860$\pm$1.350 & 7.372$\pm$4.532   \\ 
        WGWSNet \citep{zhu2023learning}          & 17.660$\pm$2.221 & 0.731$\pm$0.107 & 0.862$\pm$0.064 & 0.955$\pm$0.023 & 4.371$\pm$1.227 & 7.723$\pm$4.046   \\ 
        MvkSR \citep{xu2024mvksr}            & 31.773$\pm$2.635 & 0.941$\pm$0.025 & 0.956$\pm$0.014 & 0.989$\pm$0.004 & 3.700$\pm$0.762 & 11.185$\pm$8.973  \\ \hline
        USRNet          & \underline{34.778$\pm$4.136} 	& \underline{0.961$\pm$0.026} 	& \underline{0.975$\pm$0.012} 	& \underline{0.994$\pm$0.003} 	& 3.508$\pm$0.902 	& 6.814$\pm$2.982                    \\ \hline
        \end{tabular}\label{table:deraining}
    \end{table}
    \begin{figure*}[t]
        \centering
        \setlength{\abovecaptionskip}{0.cm}
        \includegraphics[width=1.00\linewidth]{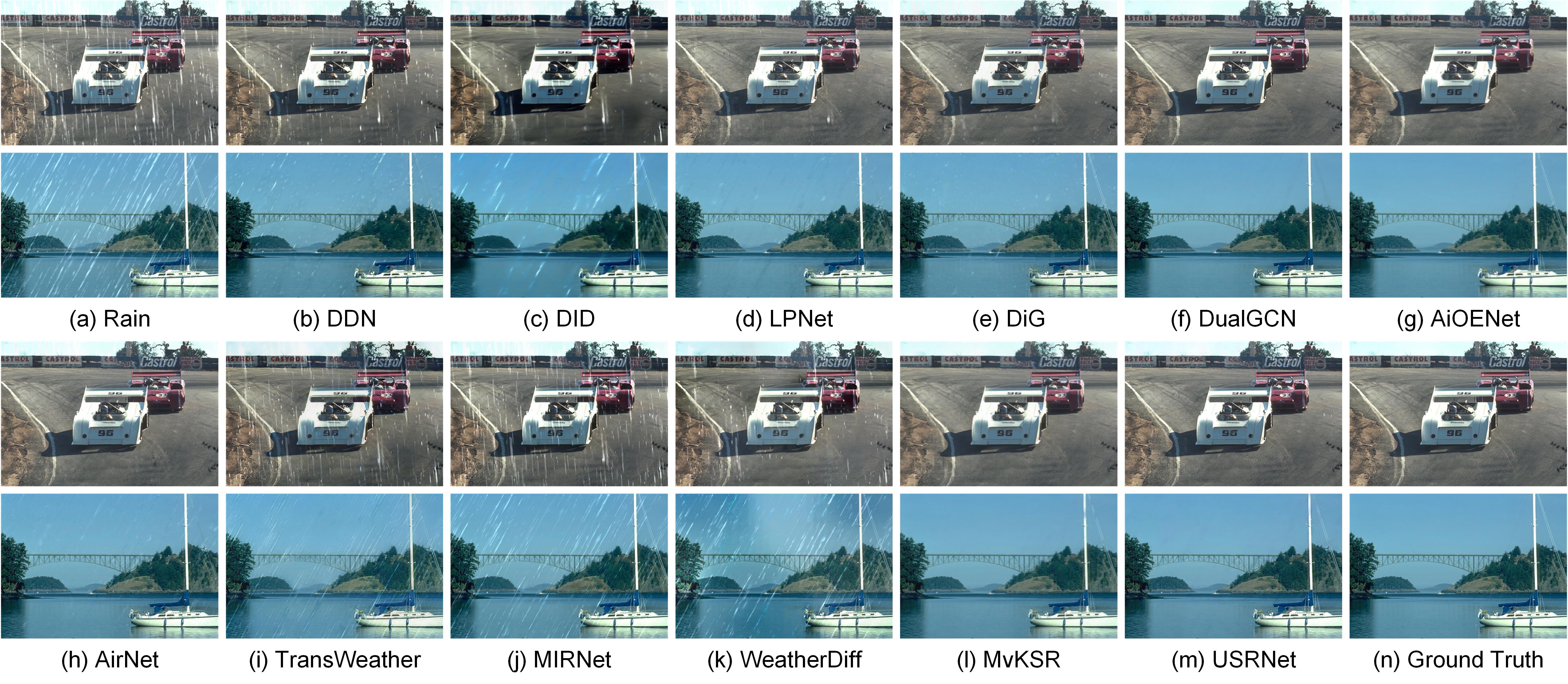}
        \caption{Visual comparisons of hazy scene recovery from RESIDE \citep{li2018benchmarking}. (a) Rain, restored images, generated by (b) DDN \citep{fu2017removing}, (c) DID \citep{zhang2018density}, (d) LPNet \citep{fu2019lightweight}, (e) DIG \citep{ran2020single}, (f) DualGCN \citep{fu2021rain}, (g) AiOENet \citep{liu2023aioenet}, (h) AirNet \citep{li2022all}, (i) TransW \citep{valanarasu2022transweather}, (j) MIRNet \citep{zamir2022learning}, (k) WeatherDiff \citep{ozdenizci2023restoring}, (l) MvkSR \citep{xu2024mvksr}, (m) USRNet, and (n) Ground Truth, respectively.}
        \label{Figure_Deraining}
    \end{figure*}
\subsubsection{Evaluation Metrics}
     To quantitatively assess the effectiveness of visibility enhancement, we utilize a variety of evaluation metrics, including both referenced and no-referenced metrics. Reference-based metrics, which require a ground truth image for comparison, include peak signal-to-noise ratio (PSNR), structural similarity index (SSIM) \citep{wang2004image}, feature similarity index (FSIM) \citep{zhang2011fsim}, and visual saliency-induced index (VSI) \citep{zhang2014vsi}. These metrics measure the fidelity of the enhanced image in terms of signal clarity and structural similarity, with higher values indicating superior quality. No-reference metrics, which do not require a reference image, include the natural image quality evaluator (NIQE) \citep{mittal2012making} and the perceptual image quality evaluator (PIQE) \citep{venkatanath2015blind}. These metrics assess the perceptual quality based on intrinsic image features, with lower values indicating better quality. Superior scene recovery is indicated by higher PSNR, SSIM, FSIM, and VSI values, as well as lower NIQE and PIQE values. These metrics provide a robust framework for comparing our method against other advanced methods, ensuring high-quality, perceptually pleasing enhancements that align with human visual preferences.
\subsubsection{Competitive Methods}
    To evaluate the restoration performance, we will compare the USRNet with several state-of-the-art methods, including single- and multi-scene methods. The dehazing methods include DCP \citep{he2010single}, MSCNN \citep{ren2016single}, AODNet \citep{li2017aod}, FFANet \citep{qin2020ffa}, DehazeFormer \citep{song2023vision}, and AiOENet \citep{liu2023aioenet}. The deraining methods include DDN \citep{fu2017removing}, DID \citep{zhang2018density}, LPNet \citep{fu2019lightweight}, DIG \citep{ran2020single}, KBNet \citep{zhang2023kbnet}, and AiOENet \citep{liu2023aioenet}. The desnowing methods include CodeNet \citep{yu2021single}, DRT \citep{liang2022drt}, SnowFormer \citep{chen2022snowformer}, UMWT \citep{kulkarni2023unified}, FocalNet \citep{cui2023focal}, and AiOENet \citep{liu2023aioenet}. The comparative AiOENet can't restore images that are mixed with multiple degradations. The multi-scene image restoration methods include AirNet \citep{li2022all}, TransWeather (TransW) \citep{valanarasu2022transweather}, MIRNet \citep{zamir2022learning}, WeatherDiffusion (WeatherDiff) \citep{ozdenizci2023restoring}, and MvKSR \citep{xu2024mvksr}. To ensure the fairness and impartiality of the experiment, all code is derived from the source files published by the author.
\subsection{Synthetic Degradation Analysis}
   In this subsection, we utilize USRNet and other competitive methods to enhance five types of low-visibility images: haze, rain, snow, haze combined with rain, and haze combined with snow. We will conduct both quantitative and qualitative analyses to evaluate the enhancement results.
    \setlength{\tabcolsep}{0.50pt}
    \begin{table}[t]
        \centering
        \scriptsize
        \caption{Comparison of desnowing quantitative results (mean$\pm$std) with referenced and no-referenced evaluation metrics on CSD \citep {chen2021all}. The best results are in \textbf{bold}, and the second best are with \underline{underline}.}
        \begin{tabular}{l|cccc|cc}
        \hline
        & PSNR $\uparrow$ & SSIM $\uparrow$ & FSIM $\uparrow$ & VSI $\uparrow$ & NIQE $\downarrow$ & PIQE $\downarrow$\\ \hline\hline
        CodeNet \citep{yu2021single}      & 21.187$\pm$2.192 & 0.713$\pm$0.048 & 0.840$\pm$0.027 & 0.955$\pm$0.013 & 5.271$\pm$0.815 & 47.564$\pm$5.401 \\ 
        DRT \citep{liang2022drt}          & 18.237$\pm$2.173 & 0.429$\pm$0.082 & 0.729$\pm$0.045 & 0.917$\pm$0.020 & 4.308$\pm$0.664 & 29.231$\pm$8.383 \\ 
        SnowFormer \citep{chen2022snowformer}   & 20.194$\pm$1.517 & 0.638$\pm$0.088 & 0.830$\pm$0.044 & 0.942$\pm$0.020 & 4.701$\pm$0.829 & 26.668$\pm$9.205 \\ 
        UMWT \citep{kulkarni2023unified}         & 20.586$\pm$2.006 & 0.668$\pm$0.082 & 0.839$\pm$0.045 & 0.945$\pm$0.020 & 4.480$\pm$0.700 & 24.714$\pm$8.360 \\ 
        FocalNet \citep{cui2023focal}     & 19.672$\pm$1.779 & 0.624$\pm$0.106 & 0.830$\pm$0.054 & 0.938$\pm$0.026 & 3.778$\pm$0.790 & 10.734$\pm$6.179 \\ 
        AiOENet \citep{liu2023aioenet}      & \underline{29.945$\pm$2.827} & \underline{0.923$\pm$0.018} & \underline{0.955$\pm$0.012} & \underline{0.989$\pm$0.004} & 3.911$\pm$0.689 & 23.076$\pm$9.154 \\ 
        AirNet \citep{li2022all}              &     19.016$\pm$2.056  & 	0.572$\pm$0.109  & 	0.803$\pm$0.047  & 	0.935$\pm$0.021 & 	4.374$\pm$1.015 & 	8.986$\pm$5.480   \\ 
        TransW \citep{valanarasu2022transweather} & 21.449$\pm$2.217 & 0.690$\pm$0.070 & 0.857$\pm$0.027 & 0.959$\pm$0.011 & 4.058$\pm$1.046 & \textbf{2.984$\pm$1.198}  \\ 
        WeatherDiff \citep{ozdenizci2023restoring}  & 22.079$\pm$2.426 & 0.755$\pm$0.083 & 0.866$\pm$0.042 & 0.955$\pm$0.017 & 3.652$\pm$0.810 & 14.477$\pm$8.614 \\ 
        WGWSNet \citep{zhu2023learning}      & 16.009$\pm$1.239 & 0.563$\pm$0.091 & 0.772$\pm$0.049 & 0.918$\pm$0.018 & 4.241$\pm$0.971 & 10.743$\pm$6.443 \\ 
        MvkSR \citep{xu2024mvksr}        &   27.868$\pm$2.447 &	0.903$\pm$0.028	 & 0.946$\pm$0.017	 &0.987$\pm$0.005	 &\underline{3.220$\pm$0.761}	 &8.224$\pm$6.761\\ \hline
        USRNet       &\textbf{30.096$\pm$2.680} 	&\textbf{0.921$\pm$0.018} 	&\textbf{0.955$\pm$0.012} 	&\textbf{0.990$\pm$0.004} 	&\textbf{3.177$\pm$0.723}	&\underline{7.507$\pm$4.428}                  \\ \hline
        \end{tabular}\label{table:desnowing}
    \end{table}
    \begin{figure*}[t]
        \centering
        \setlength{\abovecaptionskip}{0.cm}
        \includegraphics[width=1.00\linewidth]{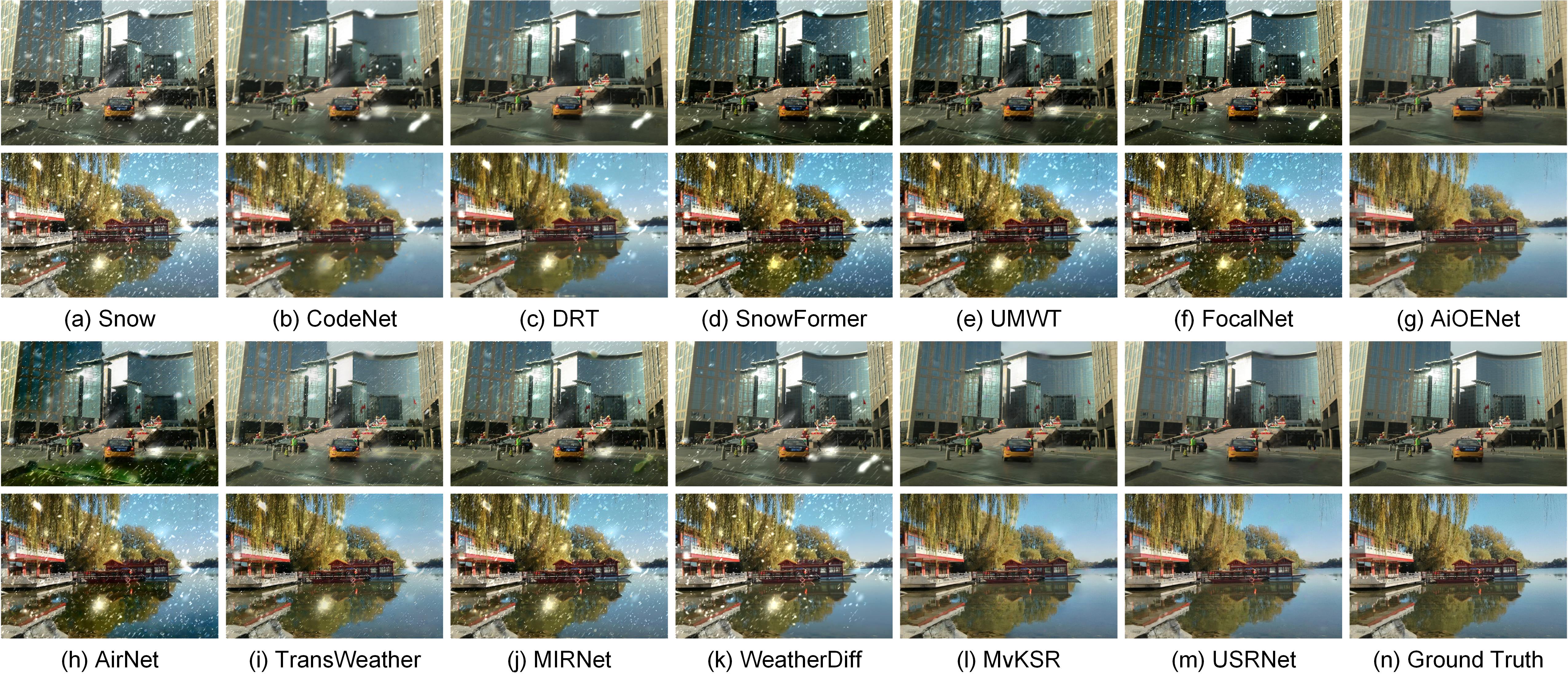}
        \caption{Visual comparisons of hazy scene recovery from CSD \citep{chen2021all}. (a) Snow, restored images, generated by (b) CodeNet \citep{yu2021single}, (c) DRT \citep{liang2022drt}, (d) SnowFormer \citep{chen2022snowformer}, (e) UMWT \citep{kulkarni2023unified}, (f) FocalNet \citep{cui2023focal}, (g) AiOENet \citep{liu2023aioenet}, (h) AirNet \citep{li2022all}, (i) TransW \citep{valanarasu2022transweather}, (j) MIRNet \citep{zamir2022learning}, (k) WeatherDiff \citep{ozdenizci2023restoring}, (l) MvkSR \citep{xu2024mvksr}, (m) USRNet, and (n) Ground Truth, respectively.}
        \label{Figure_Desnowing}
    \end{figure*}
\subsubsection{Dehazing}
    The RESIDE-OTS dataset is used for both quantitative and qualitative evaluation of dehazing methods. Table \ref{table:dehazing} indicates that physical prior-based DCP demonstrate robustness and adaptability across various hazy scenes. Most of learning-based methods, however, show instability and inferior performance, indicating their reliance on extensive and diverse training data. USRNet with its NILM, learns and generalizes features from multiple low-visibility scenes at a fine scale, surpassing limitations of single-scene feature maps. It leads to comparable enhancement results in terms of stability and metrics. Visual comparisons in Fig. \ref{Figure_Dehazing} reveal that DCP has issues with unnatural black patches, especially in sky and water area, due to transmission map estimation errors. MSCNN, AODNet, and FFANet excel in low-density haze but struggle with generalization. DehazeFormer and AirNet, though proficient in feature extraction, can introduce local distortions. TransW, MIRNet, and WeatherDiff, despite their capacity to handle dense haze, produce images with compromised contrast. AiOENet and MvKSR have a closer performance to USRNet, but its visual performance is poor in local areas with thick haze. In contrast, our proposed method distinguishes itself by delivering the most visually natural and artifact-free results across varying degrees of haze, demonstrating better adaptability and effectiveness.
\subsubsection{Deraining}
    In our deraining experiments, the Rain100L dataset is used to quantitatively evaluate different methods, focusing on advanced learning-driven methods. As shown in Table \ref{table:deraining} and Fig. \ref{Figure_Deraining}, DDN as an early method, still demonstrates satisfactory performance. DIG effectively removes rain streaks by integrating gradient detection and deep learning, but it struggles with varying rain directions. DualGCN uses graph neural networks to improve global feature extraction, yet local rain marks still impact the restored images. LPNet's lightweight design limits its ability to handle complex rain conditions. AirNet's reliance on the Rain100L dataset, despite its use of attention mechanisms, restricts its generalization capabilities. TransW and WeatherDiff excel in scenes with less rain interference, but has limitations. AiOENet and MvKSR can specifically learn the rain degradation process and have good quantitative and qualitative results. Limited by the differences in the features of the degraded images learned, MIRNet is difficult to achieve satisfactory performance in Rain100L, which further reflects the dependence of learning-based methods on training data. In contrast, our method consistently yields better quantitative results, effectively removing rain streaks of diverse shapes and directions, producing images closer to the underlying sharpness, thereby enhancing visual perception for intelligent marine vehicles in rainy conditions.
    \setlength{\tabcolsep}{0.50pt}
    \begin{table}[t]
        \centering
        \scriptsize
        \caption{Comparison of multi-scene enhancement quantitative results (mean$\pm$std) with referenced and no-referenced evaluation metrics on CDD-11 \citep{guo2024onerestore}. The best results are in \textbf{bold}, and the second best are with \underline{underline}.}
        \begin{tabular}{l|cccc|cc}
        \hline
        & PSNR $\uparrow$ & SSIM $\uparrow$ & FSIM $\uparrow$ & VSI $\uparrow$ & NIQE $\downarrow$ & PIQE $\downarrow$\\ \hline\hline
        AirNet \citep{li2022all}       & 24.644$\pm$3.266 & 0.878$\pm$0.065 & 0.948$\pm$0.025 & 0.985$\pm$0.007 & 4.911$\pm$1.034 & 9.853$\pm$4.046  \\ 
        TransW \citep{valanarasu2022transweather} & 24.398$\pm$3.118 & 0.911$\pm$0.047 & 0.956$\pm$0.024 & 0.987$\pm$0.007 & 3.969$\pm$0.726 & 7.766$\pm$3.112  \\ 
        WeatherDiff \citep{ozdenizci2023restoring}  & 23.009$\pm$2.615 & 0.918$\pm$0.041 & 0.959$\pm$0.019 & 0.988$\pm$0.006 &\textbf{ 2.624$\pm$0.572} &\textbf{ 6.700$\pm$3.390}  \\ 
        WGWSNet \citep{zhu2023learning}      & \underline{30.106$\pm$4.488} & \underline{0.961$\pm$0.025} & \underline{0.981$\pm$0.012} & \underline{0.995$\pm$0.003} & \underline{5.128$\pm$1.158} & 10.356$\pm$4.210 \\ 
        MvkSR \citep{xu2024mvksr}        &   27.639$\pm$3.791	&0.951$\pm$0.034	&0.977$\pm$0.016	&0.994$\pm$0.004	&5.193$\pm$1.154	&10.796$\pm$4.293  \\ \hline
        USRNet       &\textbf{30.756$\pm$3.987} 	&\textbf{0.964$\pm$0.033}	&\textbf{0.985$\pm$0.016} 	&\textbf{0.996$\pm$0.004} 	&3.354$\pm$0.722	&\underline{7.638$\pm$3.302}                   \\ \hline
        \end{tabular}\label{table:mix}
    \end{table}
    \begin{figure*}[t]
        \centering
        \setlength{\abovecaptionskip}{0.cm}
        \includegraphics[width=1.00\linewidth]{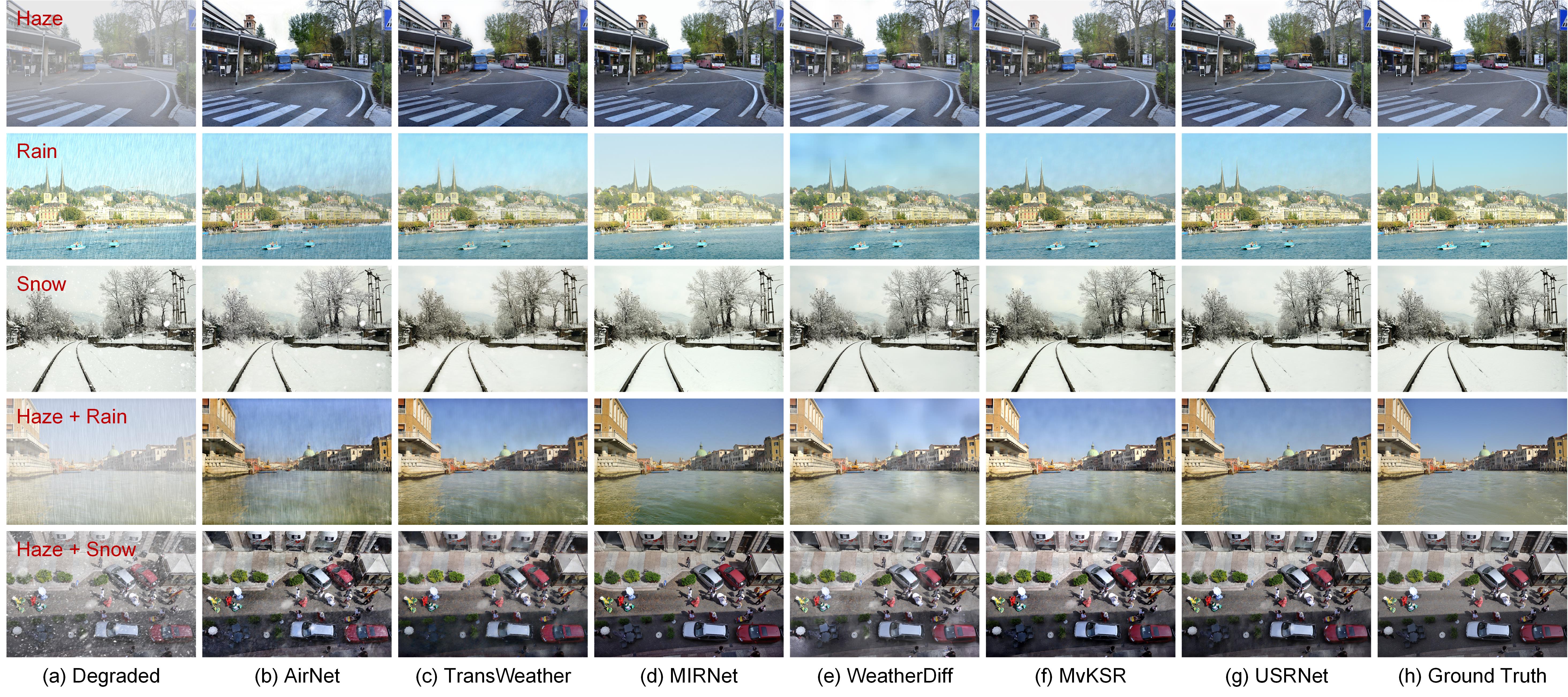}
        \caption{Visual comparisons of hazy scene recovery from CDD-11 \citep{guo2024onerestore}. (a) Degraded, restored images, generated by (b) AirNet \citep{li2022all}, (c) TransW \citep{valanarasu2022transweather}, (d) MIRNet \citep{zamir2022learning}, (e) WeatherDiff \citep{ozdenizci2023restoring}, (f) MvkSR \citep{xu2024mvksr}, (g) USRNet, and (h) Ground Truth, respectively.}
        \label{Figure_CDD}
    \end{figure*}
\subsubsection{Desnowing}
    \begin{figure*}[t]
        \centering
        \includegraphics[width=1.00\linewidth]{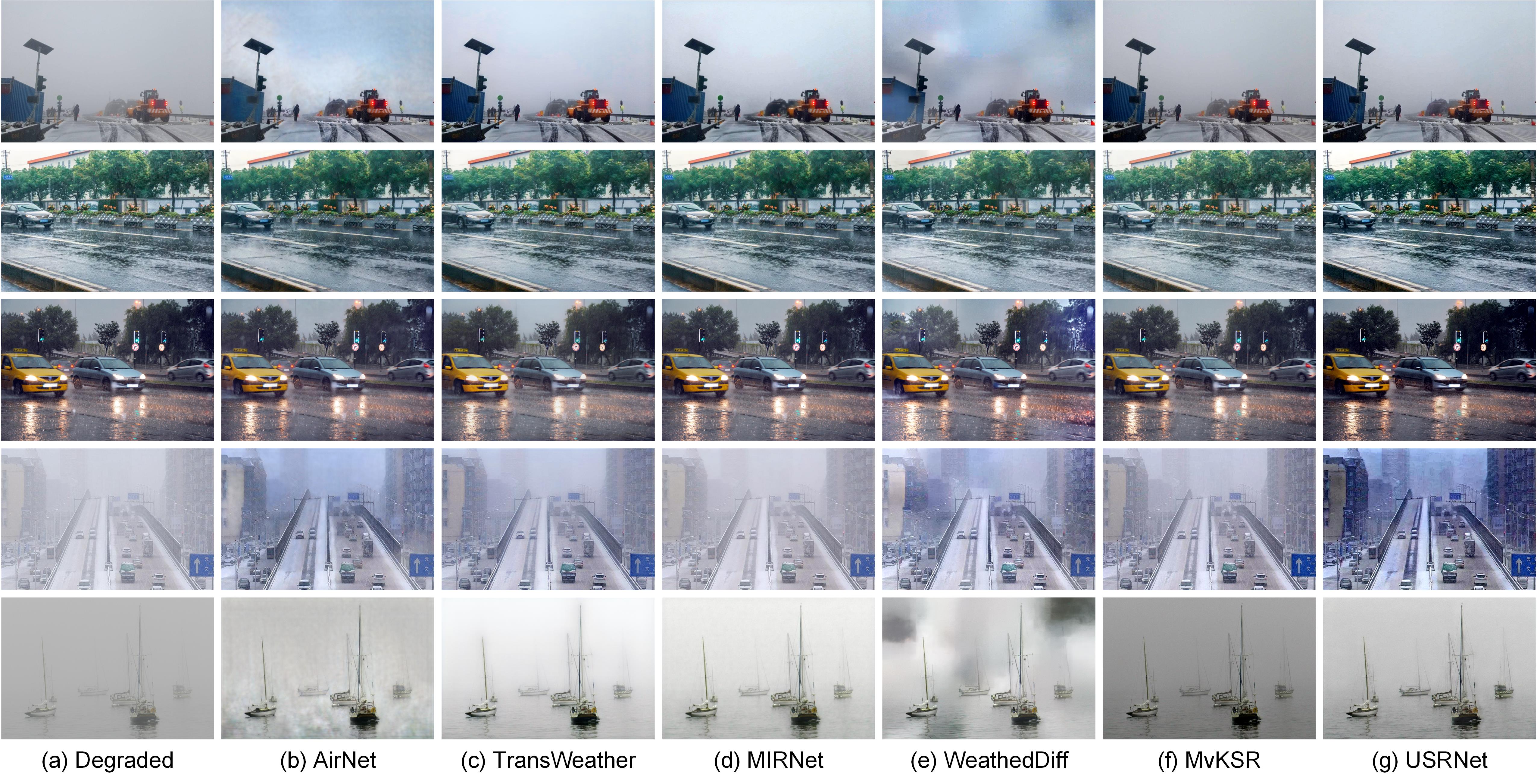}
        \caption{Visual comparisons of scene recovery performance from real-world low-visibility images. (a) Real-world low-visibility images, restored images, generated by (b) AirNet \citep{li2022all}, (c) TransWeather \citep{valanarasu2022transweather}, (d) MIRNet \citep{zamir2020learning}, (e) WeatherDiff \citep{ozdenizci2023restoring}, (f) MvKSR \citep{xu2024mvksr}, and (g) USRNet, respectively.}
        \label{Fig_real}
    \end{figure*}
    Image desnowing presents a more complex challenge compared to rain removal, primarily due to the extensive occlusions caused by snowflakes. To assess the performance of various desnowing techniques, we used synthetic snowy images from the CSD dataset. As detailed in Table \ref{table:desnowing}, a comparison of desnowing methods reveals nuanced strengths and weaknesses. CODENet performs well with small snow marks, while DRT, a lightweight network, struggles with large-scale snow. The models SnowFormer, UMWT, and FocalNet, despite utilizing attention mechanisms, lack robustness across different scenes. WeatherDiff is limited to small-scale snow. AirNet and TransW are insufficiently robust for diverse desnowing tasks. In Fig. \ref{Figure_Desnowing}, snow patches degrade visual quality, especially when covering vessel targets. CODENet handles small snow but not large-scale coverage. DRT is efficient but limited. SnowFormer, UMWT, and FocalNet struggle to separate snow from the background. A noteworthy observation is that even restoration networks designed for general scenes yield comparable quantitative results, implying a shared limitation, that is, deep networks' performance is tightly coupled with the diversity and specificity of their training datasets. Our USRNet demonstrates a distinct advantage in reconstructing areas heavily affected by large-scale snowfall, thereby significantly improving image quality and visual perception, highlighting its robustness and adaptability in extreme weather image restoration.

\subsubsection{Mixed Degradation Restoration}
    Similar to the dehazing, deraining, and desnowing tasks, we first quantify the performance of our method on mixed degraded images from the CDD-11 \citep{guo2024onerestore} by employing a suite of objective evaluation metrics. As detailed in Table \ref{table:mix}, USRNet consistently outperforms competitors across all metrics, highlighting its superiority. The recovery of scenes contaminated by both haze and rain/snow is a notably complex endeavor. A visual inspection of Fig. \ref{Figure_CDD}, reveals the limitations of existing techniques in handling these compound conditions. Conversely, USRNet demonstrates impressive resilience, delivering superior results in the challenging task of recovering images with complex mixed degradation, thereby showcasing its adaptability to complex mixed scene restorations.
    \setlength{\tabcolsep}{1.6pt}
    \begin{table}[t]
        \centering
        \scriptsize
        \caption{Ablation analysis (PSNR / SSIM) of the suggested module on CDD-11 \citep{guo2024onerestore}.}
        \begin{tabular}{cc|cccccc}
        \hline
        One-to-One  & All-in-One                       & Haze           & Rain   & Snow & Haze + Rain & Haze+ Snow    & Average      \\ \hline\hline
        \CheckmarkBold &  & 29.302/0.984 & 32.665/0.957 & 32.681/0.952 & 28.113/0.948 & 28.041/0.932 & 30.160/0.955\\\hline
        & \CheckmarkBold & 29.383/0.983 & 32.600/0.956 & 32.627/0.951 & 27.639/0.944 & 27.797/0.932 & 30.009/0.953\\ \hline
        \end{tabular}\label{table:one-all}
    \end{table}
    \setlength{\tabcolsep}{11.00pt}
    \begin{table}[t]
        \centering
        \scriptsize
        \caption{Ablation analysis (PSNR and SSIM) of the suggested D-Res on CDD-11 \citep{guo2024onerestore}.}
        \begin{tabular}{c|ccc|cc}
        \hline
        \multirow{2}{*}{SCL}          & \multicolumn{3}{c|}{DCL}                                                                                                                & \multirow{2}{*}{PSNR $\uparrow$} & \multirow{2}{*}{SSIM $\uparrow$}  \\ \cline{2-4} & $K_s$ & $K_d$  & $K_L$  &  &    \\ \hline\hline
        \CheckmarkBold &                               &                               &                & 28.831 & 0.933  \\
        \CheckmarkBold & \CheckmarkBold &                               &                               & 29.335 & 0.939  \\
        \CheckmarkBold &                               & \CheckmarkBold &                               & 29.697 & 0.947 \\\hline
        \CheckmarkBold &                               & \CheckmarkBold & \CheckmarkBold & 30.009 & 0.953 \\ \hline
        \end{tabular}\label{table:dres}
    \end{table}
\subsection{Real-world Degradation Analysis}
    The real-world low-visibility imaging process in VITS is more complicated. As shown in Fig. \ref{Fig_real}, we selected mixed degraded images related to land/ocean for visual comparison. AirNet exhibits color distortion in local areas accompanied by unnatural gradient changes. TransW and MIRNet perform well overall, especially in the reconstruction of global contrast, but cannot completely eliminate the degradation effect when the haze density is too high. The image restored by WeatherDiff has local artifacts. MvKSR also faces the problem of difficulty in accurately extracting features from water fog. Thanks to the sub-node training mechanism of USRNet, the challenges of different types of degradation can be more naturally addressed, thus achieving the best visual performance.
    \begin{figure*}[t]
        \centering
        \setlength{\abovecaptionskip}{0.cm}
        \includegraphics[width=1.00\linewidth]{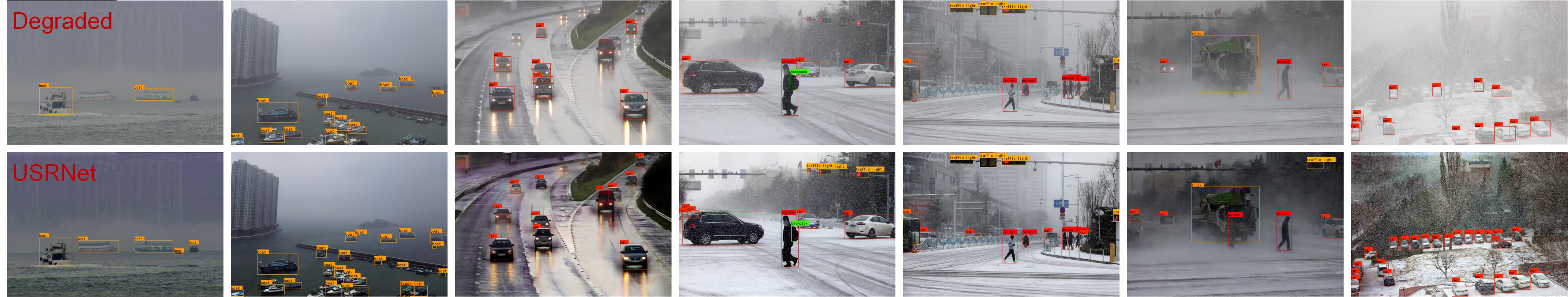}
        \caption{Comparisons of YOLOv10-based object detection results for visually-degraded images and USRNet-restored versions.}
        \label{Figure_Dete}
    \end{figure*}
    \setlength{\tabcolsep}{2.00pt}
    \begin{table}[t]
        \centering
        \scriptsize
        \caption{Detection average AP of YOLOv10 on the transportation-related synthetic sandstorm images from VOC and the images enhanced by various methods.}
       \begin{tabular}{l|cccccccc}
        \hline
        Methods                       & Aer.        & Bic.          & Boat             & Bus              & Car              & Mot.        & Person                       & AP              \\ \hline \hline
        AirNet \citep{li2022all}                         & 0.881 & 0.931 & 0.715 & 0.948 & 0.865 & 0.945 & 0.910 & 0.844  \\ 
        TransW \citep{valanarasu2022transweather}                         & 0.885 & 0.935 & 0.725 & 0.958 & 0.868 & 0.947 & 0.914 & 0.847 \\ 
        MIRNet \citep{zamir2022learning}                         & 0.901 & 0.937 & 0.715 & 0.941 & 0.866 & 0.938 & 0.906 & 0.851 \\ 
        WeatherDiff \citep{ozdenizci2023restoring}                         & 0.892 & 0.927 & 0.711 & 0.937 & 0.871 & 0.938 & 0.898 & 0.841  \\ 
        MvKSR \citep{xu2024mvksr}                         & 0.939 & 0.954 & 0.771 & 0.969 & 0.869 & 0.957 & 0.925 & 0.885 \\ 
        USRNet                         & 0.969 & 0.958 & 0.794 & 0.971 & 0.865 & 0.960 & 0.927 & 0.891 \\ \hline
        Ground Truth                   & 0.976 & 0.963     & 0.819     & 0.975 & 0.859 & 0.961  & 0.930 & 0.922  \\ \hline
        \end{tabular}
        \label{Table_detection}
    \end{table}
\subsection{Ablation Study}\label{as}
    Ablation study is an effective method to analyze which modules play a key role in network learning. Therefore, we will conduct ablation studies from two aspects: NILM and D-Res to more accurately analyze the key parts.
\subsubsection{Ablation Studies on NILM}
    Table \ref{table:one-all} shows the restoration performance of our method in one-to-one (i.e., only retaining the learning nodes for specific degradation in the reasoning stage) and many-to-one. Obviously, our joint reasoning of multiple tasks will not significantly reduce the quantitative indicators, and no human intervention is required to achieve robust multi-scene image restoration tasks. 
\subsubsection{Ablation Studies on D-Res}
    As a crucial component for extracting edges and global features, D-Res was disassembled and retrained under identical conditions. As demonstrated in Table \ref{table:dres}, dilated convolution achieves higher quantitative index values compared to standard convolution. This improvement is primarily because dilated convolution can more effectively restore damaged edges and other features through its larger receptive field. Furthermore, incorporating the Laplacian operator enhances this capability even further.
\subsection{Improving High-level Tasks with USRNet}
    We tackle the challenging problem of object detection and recognition in diverse weather conditions, shedding light on the intricate interplay between high-level vision tasks and image restoration. As a robust baseline, we adopt the YOLOV10 \citep{wang2024yolov10} and assess its performance on the VOC dataset using both synthetically degraded and naturally degraded images. Our experimental results uncover a significant trend: as image degradation intensifies and becomes more complex, object detection reliability drops substantially. Notably, our proposed method consistently outperforms other approaches in enhancing the detection accuracy of YOLOV10 in adverse weather conditions, as demonstrated in Table \ref{Table_detection}. Fig. \ref{Figure_Dete} shows the restoration effect of our method compared to the original degraded image, This underscores the effectiveness of our method in mitigating the adverse effects of degradation on object detection, thereby improving the robustness of vision systems in real-world applications.

\section{Conclusion}\label{sec:conclusion}
    This work presents a novel unified scene restoration network (USRNet) aimed at addressing complex imaging scene degradation in VITS. USRNet comprises a scene encoder, an attention-driven NILM, an edge decoder, and a scene restorer. The scene encoder extracts deep features through advanced residual blocks, ensuring comprehensive degradation encoding. NILM innovatively allows the network to independently learn and respond to different weather conditions, thereby enhancing its adaptability and robustness. The edge decoder meticulously extracts edge features, crucial for maintaining image sharpness. In addition, we propose a hybrid loss function to tune the training to extraction various degradation nuances. Extensive experimental results demonstrate USRNet's superiority in handling complex image degradation scenarios.
\bibliographystyle{elsarticle-num}
\footnotesize
\bibliography{USRNet}

\begin{thebibliography}{10}
\expandafter\ifx\csname url\endcsname\relax
  \def\url#1{\texttt{#1}}\fi
\expandafter\ifx\csname urlprefix\endcsname\relax\def\urlprefix{URL }\fi
\expandafter\ifx\csname href\endcsname\relax
  \def\href#1#2{#2} \def\path#1{#1}\fi

\bibitem{wan2022edge}
S.~Wan, S.~Ding, C.~Chen, Edge computing enabled video segmentation for real-time traffic monitoring in internet of vehicles, Pattern Recognit. 121 (2022) 108146.

\bibitem{liu2023aioenet}
R.~W. Liu, Y.~Lu, Y.~Guo, W.~Ren, F.~Zhu, Y.~Lv, Aioenet: All-in-one low-visibility enhancement to improve visual perception for intelligent marine vehicles under severe weather conditions, IEEE Trans. Intell. Veh. (2023).

\bibitem{husain2020vehicle}
A.~A. Husain, T.~Maity, R.~K. Yadav, Vehicle detection in intelligent transport system under a hazy environment: a survey, IET Image Proc. 14~(1) (2020) 1--10.

\bibitem{he2010single}
K.~He, J.~Sun, X.~Tang, Single image haze removal using dark channel prior, IEEE Trans. Pattern Anal. Mach. Intell. 33~(12) (2010) 2341--2353.

\bibitem{chen2024dea}
Z.~Chen, Z.~He, Z.-M. Lu, Dea-net: Single image dehazing based on detail-enhanced convolution and content-guided attention, IEEE Trans. Image Process. (2024).

\bibitem{fu2017clearing}
X.~Fu, J.~Huang, X.~Ding, Y.~Liao, J.~Paisley, Clearing the skies: A deep network architecture for single-image rain removal, IEEE Trans. Image Process. 26~(6) (2017) 2944--2956.

\bibitem{liu2018desnownet}
Y.-F. Liu, D.-W. Jaw, S.-C. Huang, J.-N. Hwang, Desnownet: Context-aware deep network for snow removal, IEEE Trans. Image Process. 27~(6) (2018) 3064--3073.

\bibitem{zhu2023learning}
Y.~Zhu, T.~Wang, X.~Fu, X.~Yang, X.~Guo, J.~Dai, Y.~Qiao, X.~Hu, Learning weather-general and weather-specific features for image restoration under multiple adverse weather conditions, in: Proc. IEEE CVPR, 2023, pp. 21747--21758.

\bibitem{liu2024residual}
H.~Liu, A.~Zhang, W.~Zhu, B.~Fu, B.~Ding, S.~Xiong, Residual deformable convolution for better image de-weathering, Pattern Recognit. 147 (2024) 110093.

\bibitem{xu2024mvksr}
W.~Xu, D.~Yang, Y.~Gao, Y.~Lu, J.~Zhang, Y.~Guo, Mvksr: Multi-view knowledge-guided scene recovery for hazy and rainy degradation, IEEE Trans. Instrum. Meas. (2024).

\bibitem{guo2024onerestore}
Y.~Guo, Y.~Gao, Y.~Lu, R.~W. Liu, S.~He, Onerestore: A universal restoration framework for composite degradation, in: Proc. ECCV, 2024, pp. 0--1.

\bibitem{fu2014retinex}
X.~Fu, P.~Zhuang, Y.~Huang, Y.~Liao, X.-P. Zhang, X.~Ding, A retinex-based enhancing approach for single underwater image, in: Proc. IEEE ICIP, 2014, pp. 4572--4576.

\bibitem{ren2016single}
W.~Ren, S.~Liu, H.~Zhang, J.~Pan, X.~Cao, M.-H. Yang, Single image dehazing via multi-scale convolutional neural networks, in: Proc. ECCV, 2016, pp. 154--169.

\bibitem{song2023vision}
Y.~Song, Z.~He, H.~Qian, X.~Du, Vision transformers for single image dehazing, IEEE Trans. Image Process. 32 (2023) 1927--1941.

\bibitem{guo2023scanet}
Y.~Guo, Y.~Gao, W.~Liu, Y.~Lu, J.~Qu, S.~He, W.~Ren, Scanet: Self-paced semi-curricular attention network for non-homogeneous image dehazing, in: Proc. IEEE CVPRW, 2023, pp. 1884--1893.

\bibitem{liu2024dfp}
J.~Liu, S.~Wang, C.~Chen, Q.~Hou, Dfp-net: An unsupervised dual-branch frequency-domain processing framework for single image dehazing, Eng. Appl. Artif. Intell. 136 (2024) 109012.

\bibitem{xu2012removing}
J.~Xu, W.~Zhao, P.~Liu, X.~Tang, Removing rain and snow in a single image using guided filter, in: Proc. IEEE CSAE, 2012, pp. 304--307.

\bibitem{wang2017hierarchical}
Y.~Wang, S.~Liu, C.~Chen, B.~Zeng, A hierarchical approach for rain or snow removing in a single color image, IEEE Trans. Image Process. 26~(8) (2017) 3936--3950.

\bibitem{quan2023image}
Y.~Quan, X.~Tan, Y.~Huang, Y.~Xu, H.~Ji, Image desnowing via deep invertible separation, IEEE Trans. Circuits Syst. Video Technol. 33~(7) (2023) 3133--3144.

\bibitem{yang2022rain}
F.~Yang, J.~Ren, Z.~Lu, J.~Zhang, Q.~Zhang, Rain-component-aware capsule-gan for single image de-raining, Pattern Recognit. 123 (2022) 108377.

\bibitem{liu2022tape}
L.~Liu, L.~Xie, X.~Zhang, S.~Yuan, X.~Chen, W.~Zhou, H.~Li, Q.~Tian, Tape: Task-agnostic prior embedding for image restoration, in: Proc. ECCV, Springer, 2022, pp. 447--464.

\bibitem{guo2020joint}
Y.~Guo, J.~Chen, X.~Ren, A.~Wang, W.~Wang, Joint raindrop and haze removal from a single image, IEEE Trans. Image Process. 29 (2020) 9508--9519.

\bibitem{cheng2023highway}
X.~Cheng, J.~Zhou, J.~Song, X.~Zhao, A highway traffic image enhancement algorithm based on improved gan in complex weather conditions, IEEE Trans. Intell. Transp. Syst. 24~(8) (2023) 8716--8726.

\bibitem{chen2021pre}
H.~Chen, Y.~Wang, T.~Guo, C.~Xu, Y.~Deng, Z.~Liu, S.~Ma, C.~Xu, C.~Xu, W.~Gao, Pre-trained image processing transformer, in: Proc. IEEE CVPR, 2021, pp. 12299--12310.

\bibitem{ozdenizci2023restoring}
O.~{\"O}zdenizci, R.~Legenstein, Restoring vision in adverse weather conditions with patch-based denoising diffusion models, IEEE Trans. Pattern Anal. Mach. Intell. (Jan. 2023).

\bibitem{ye2024learning}
T.~Ye, S.~Chen, W.~Chai, Z.~Xing, J.~Qin, G.~Lin, L.~Zhu, Learning diffusion texture priors for image restoration, in: Proc. IEEE CVPR, 2024, pp. 2524--2534.

\bibitem{lin2024improving}
J.~Lin, Z.~Zhang, Y.~Wei, D.~Ren, D.~Jiang, Q.~Tian, W.~Zuo, Improving image restoration through removing degradations in textual representations, in: Proc. IEEE CVPR, 2024, pp. 2866--2878.

\bibitem{fattal2014dehazing}
R.~Fattal, Dehazing using color-lines, ACM Trans. Graphics 34~(1) (2014) 1--14.

\bibitem{liu2022rank}
J.~Liu, R.~W. Liu, J.~Sun, T.~Zeng, Rank-one prior: Real-time scene recovery, IEEE Trans. Pattern Anal. Mach. Intell. 45~(7) (2022) 8845--8860.

\bibitem{kandhway2023adaptive}
P.~Kandhway, An adaptive low-light image enhancement using canonical correlation analysis, IEEE Trans. Ind. Inf. (2023).

\bibitem{zhu2023spectral}
Z.~Zhu, D.~Zhang, Z.~Wang, S.~Feng, P.~Duan, Spectral dual-channel encoding for image dehazing, IEEE Trans. Circuits Syst. Video Technol. (2023).

\bibitem{zhao2024cycle}
C.~Zhao, W.~Cai, C.~Hu, Z.~Yuan, Cycle contrastive adversarial learning with structural consistency for unsupervised high-quality image deraining transformer, Neural Networks (2024) 106428.

\bibitem{yang2024single}
Y.~Yang, Y.~Zhang, Z.~Cui, H.~Zhao, T.~Ouyang, Single image deraining using scale constraint iterative update network, Expert Syst. Appl. 236 (2024) 121339.

\bibitem{zhang2021deep}
K.~Zhang, R.~Li, Y.~Yu, W.~Luo, C.~Li, Deep dense multi-scale network for snow removal using semantic and depth priors, IEEE Trans. Image Process. 30 (2021) 7419--7431.

\bibitem{chen2020jstasr}
W.-T. Chen, H.-Y. Fang, J.-J. Ding, C.-C. Tsai, S.-Y. Kuo, Jstasr: Joint size and transparency-aware snow removal algorithm based on modified partial convolution and veiling effect removal, in: Proc. ECCV, Springer, 2020, pp. 754--770.

\bibitem{chen2021all}
W.-T. Chen, H.-Y. Fang, C.-L. Hsieh, C.-C. Tsai, I.~Chen, J.-J. Ding, S.-Y. Kuo, et~al., All snow removed: Single image desnowing algorithm using hierarchical dual-tree complex wavelet representation and contradict channel loss, in: Proc. IEEE ICCV, 2021, pp. 4196--4205.

\bibitem{gou2020clearer}
Y.~Gou, B.~Li, Z.~Liu, S.~Yang, X.~Peng, Clearer: Multi-scale neural architecture search for image restoration, NeurIPS 33 (2020) 17129--17140.

\bibitem{zamir2021multi}
S.~W. Zamir, A.~Arora, S.~Khan, M.~Hayat, F.~S. Khan, M.-H. Yang, L.~Shao, Multi-stage progressive image restoration, in: Proc. IEEE CVPR, 2021, pp. 14821--14831.

\bibitem{li2020all}
R.~Li, R.~T. Tan, L.-F. Cheong, All in one bad weather removal using architectural search, in: Proc. IEEE CVPR, 2020, pp. 3175--3185.

\bibitem{li2022all}
B.~Li, X.~Liu, P.~Hu, Z.~Wu, J.~Lv, X.~Peng, All-in-one image restoration for unknown corruption, in: Proc. IEEE CVPR, 2022, pp. 17452--17462.

\bibitem{patil2023multi}
P.~W. Patil, S.~Gupta, S.~Rana, S.~Venkatesh, S.~Murala, Multi-weather image restoration via domain translation, in: Proc. IEEE ICCV, 2023, pp. 21696--21705.

\bibitem{fang2024taenet}
W.~Fang, C.~Wang, Z.~Li, A.~Grau, T.~Lai, J.~Chen, Taenet: transencoder-based all-in-one image enhancement with depth awareness, Appl. Intell. (2024) 1--22.

\bibitem{liu2024real}
R.~W. Liu, Y.~Lu, Y.~Gao, Y.~Guo, W.~Ren, F.~Zhu, F.-Y. Wang, Real-time multi-scene visibility enhancement for promoting navigational safety of vessels under complex weather conditions, IEEE Trans. Intell. Transp. Syst. (2024).

\bibitem{gao2024prompt}
H.~Gao, J.~Yang, Y.~Zhang, N.~Wang, J.~Yang, D.~Dang, Prompt-based ingredient-oriented all-in-one image restoration, IEEE Trans. Circuits Syst. Video Technol. (2024).

\bibitem{wang2022uformer}
Z.~Wang, X.~Cun, J.~Bao, W.~Zhou, J.~Liu, H.~Li, Uformer: A general u-shaped transformer for image restoration, in: Proc. IEEE CVPR, 2022, pp. 17683--17693.

\bibitem{ai2024multimodal}
Y.~Ai, H.~Huang, X.~Zhou, J.~Wang, R.~He, Multimodal prompt perceiver: Empower adaptiveness generalizability and fidelity for all-in-one image restoration, in: Proc. IEEE CVPR, 2024, pp. 25432--25444.

\bibitem{li2019heavy}
R.~Li, L.-F. Cheong, R.~T. Tan, Heavy rain image restoration: Integrating physics model and conditional adversarial learning, in: Proc. IEEE CVPR, 2019, pp. 1633--1642.

\bibitem{zhang2023data}
Z.~Zhang, Y.~Wei, H.~Zhang, Y.~Yang, S.~Yan, M.~Wang, Data-driven single image deraining: A comprehensive review and new perspectives, Pattern Recognit. 143 (2023) 109740.

\bibitem{li2018benchmarking}
B.~Li, W.~Ren, D.~Fu, D.~Tao, D.~Feng, W.~Zeng, Z.~Wang, Benchmarking single-image dehazing and beyond, IEEE Trans. Image Process. 28~(1) (2018) 492--505.

\bibitem{yang2019joint}
W.~Yang, R.~T. Tan, J.~Feng, Z.~Guo, S.~Yan, J.~Liu, Joint rain detection and removal from a single image with contextualized deep networks, IEEE Trans. Pattern Anal. Mach. Intell. 42~(6) (2019) 1377--1393.

\bibitem{wu2021contrastive}
H.~Wu, Y.~Qu, S.~Lin, J.~Zhou, R.~Qiao, Z.~Zhang, Y.~Xie, L.~Ma, Contrastive learning for compact single image dehazing, in: Proc. IEEE CVPR, 2021, pp. 10551--10560.

\bibitem{li2017aod}
B.~Li, X.~Peng, Z.~Wang, J.~Xu, D.~Feng, Aod-net: All-in-one dehazing network, in: Proc. IEEE ICCV, 2017, pp. 4770--4778.

\bibitem{qin2020ffa}
X.~Qin, Z.~Wang, Y.~Bai, X.~Xie, H.~Jia, Ffa-net: Feature fusion attention network for single image dehazing, in: Proc. AAAI, 2020, pp. 11908--11915.

\bibitem{valanarasu2022transweather}
J.~M.~J. Valanarasu, R.~Yasarla, V.~M. Patel, Transweather: Transformer-based restoration of images degraded by adverse weather conditions, in: Proc. IEEE CVPR, 2022, pp. 2353--2363.

\bibitem{zamir2022learning}
S.~W. Zamir, A.~Arora, S.~Khan, M.~Hayat, F.~S. Khan, M.-H. Yang, L.~Shao, Learning enriched features for fast image restoration and enhancement, IEEE Trans. Pattern Anal. Mach. Intell. 45~(2) (2022) 1934--1948.

\bibitem{fu2017removing}
X.~Fu, J.~Huang, D.~Zeng, Y.~Huang, X.~Ding, J.~Paisley, Removing rain from single images via a deep detail network, in: Proc. IEEE CVPR, 2017, pp. 3855--3863.

\bibitem{zhang2018density}
H.~Zhang, V.~M. Patel, Density-aware single image de-raining using a multi-stream dense network, in: Proc. IEEE CVPR, 2018, pp. 695--704.

\bibitem{fu2019lightweight}
X.~Fu, B.~Liang, Y.~Huang, X.~Ding, J.~Paisley, Lightweight pyramid networks for image deraining, IEEE Trans. Neur. Net. Lear. 31~(6) (2019) 1794--1807.

\bibitem{ran2020single}
W.~Ran, Y.~Yang, H.~Lu, Single image rain removal boosting via directional gradient, in: Proc. IEEE ICME, 2020, pp. 1--6.

\bibitem{fu2021rain}
X.~Fu, Q.~Qi, Z.-J. Zha, Y.~Zhu, X.~Ding, Rain streak removal via dual graph convolutional network, in: Proc. AAAI, 2021, pp. 1352--1360.

\bibitem{wang2004image}
Z.~Wang, A.~C. Bovik, H.~R. Sheikh, E.~P. Simoncelli, Image quality assessment: from error visibility to structural similarity, IEEE Trans. Image Process. 13~(4) (2004) 600--612.

\bibitem{zhang2011fsim}
L.~Zhang, L.~Zhang, X.~Mou, D.~Zhang, Fsim: A feature similarity index for image quality assessment, IEEE Trans. Image Process. 20~(8) (2011) 2378--2386.

\bibitem{zhang2014vsi}
L.~Zhang, Y.~Shen, H.~Li, Vsi: A visual saliency-induced index for perceptual image quality assessment, IEEE Trans. Image Process. 23~(10) (2014) 4270--4281.

\bibitem{mittal2012making}
A.~Mittal, R.~Soundararajan, A.~C. Bovik, Making a “completely blind” image quality analyzer, IEEE Signal Proc. Let. 20~(3) (2012) 209--212.

\bibitem{venkatanath2015blind}
N.~Venkatanath, D.~Praneeth, M.~C. Bh, S.~S. Channappayya, S.~S. Medasani, Blind image quality evaluation using perception based features, in: Proc. NCC, 2015, pp. 1--6.

\bibitem{zhang2023kbnet}
Y.~Zhang, D.~Li, X.~Shi, D.~He, K.~Song, X.~Wang, H.~Qin, H.~Li, Kbnet: Kernel basis network for image restoration, arXiv preprint arXiv:2303.02881 (2023).

\bibitem{yu2021single}
L.~Yu, B.~Wang, J.~He, G.-S. Xia, W.~Yang, Single image deraining with continuous rain density estimation, IEEE Trans. Multimedia 25 (2021) 443--456.

\bibitem{liang2022drt}
Y.~Liang, S.~Anwar, Y.~Liu, Drt: A lightweight single image deraining recursive transformer, in: Proc. IEEE CVPR, 2022, pp. 589--598.

\bibitem{chen2022snowformer}
S.~Chen, T.~Ye, Y.~Liu, E.~Chen, J.~Shi, J.~Zhou, Snowformer: Scale-aware transformer via context interaction for single image desnowing, arXiv preprint arXiv:2208.09703 (Aug. 2022).

\bibitem{kulkarni2023unified}
A.~Kulkarni, S.~S. Phutke, S.~Murala, Unified transformer network for multi-weather image restoration, in: Proc. ECCV, Springer, 2023, pp. 344--360.

\bibitem{cui2023focal}
Y.~Cui, W.~Ren, X.~Cao, A.~Knoll, Focal network for image restoration, in: Proc. IEEE CVPR, 2023, pp. 13001--13011.

\bibitem{zamir2020learning}
S.~W. Zamir, A.~Arora, S.~Khan, M.~Hayat, F.~S. Khan, M.-H. Yang, L.~Shao, Learning enriched features for real image restoration and enhancement, in: Proc. ECCV, 2020, pp. 492--511.

\bibitem{wang2024yolov10}
A.~Wang, H.~Chen, L.~Liu, K.~Chen, Z.~Lin, J.~Han, G.~Ding, Yolov10: Real-time end-to-end object detection, arXiv preprint arXiv:2405.14458 (2024).

\end{thebibliography}

\end{document}